\documentclass[runningheads]{llncs}

\usepackage{eccv}

\usepackage{eccvabbrv}

\usepackage[table]{xcolor}
\usepackage{adjustbox}
\usepackage{wrapfig}
\newcommand{\params}[1]{\multicolumn{2}{c|}{\small\textcolor{gray}{\textit{$#1 \times 10^5$}}}}
\newcommand{\paramslast}[1]{\multicolumn{2}{c}{\small\textcolor{gray}{\textit{$#1 \times 10^5$}}}}
\usepackage{placeins}   
\usepackage{caption}

\usepackage{arydshln}
\ADLdrawingmode{3}
\usepackage[utf8]{inputenc}
\usepackage{booktabs}
\usepackage{multirow}
\usepackage{array}
\usepackage{xcolor}
\usepackage{makecell}

\usepackage{graphicx}
\usepackage{booktabs}

\usepackage[accsupp]{axessibility}

\usepackage{xr-hyper}
\usepackage{hyperref}

\usepackage{orcidlink}

\begin{document}

\title{REAL-OW: \underline{RE}hears\underline{AL}-free \underline{O}pen \underline{W}orld Object Detection with Low-Rank Adaptation and Dual-Stage Objectness Modeling}

\titlerunning{REAL-OW}

\author{Huazhong Zhang\inst{1} \and
Yang Zhang\inst{1}\thanks{Corresponding author.} \and
Xiaowen Fu\inst{1} \and
Linlin Shen\inst{1} \and
Jinbao Wang\inst{1}
}

\authorrunning{H.~Zhang et al.}

\institute{Shenzhen University, Shenzhen, China\\
\email{yangzhang@szu.edu.cn}}

\maketitle

\begin{abstract}
Open-World Object Detection (OWOD) requires detectors to identify previously unseen objects as unknown and incrementally incorporate them into the set of known categories, while preserving previously acquired knowledge. Existing frameworks rely heavily on exemplar replay to mitigate catastrophic forgetting,
but in some real applications, storing raw data conflicts with data access restrictions and leads to data exposure risks, while incurring significant memory overhead.
In this paper, we propose REAL-OW,
a novel rehearsal-free framework that decouples incremental knowledge through a collaborative adapter architecture based on Low-Rank Adaptation (LoRA).
Specifically, we deploy General Adapters (GAs) in the backbone to enable the significance-aware refinement of cross-task universal representations, while Specific Adapters (SAs) in the decoder provide orthogonal storage for task-specific expertise. To resolve representation drift in objectness modeling under rehearsal-free constraints, we introduce Dual-Stage Objectness Modeling (DSOM), which alternates between feature aggregation and boundary consolidation to stabilize objectness distributions while maintaining the separation between known and unknown categories.
Furthermore, DSOM is supported by a Calibrated Gaussian Negative Log-Likelihood (CG-NLL) distance tailored for the dispersed feature distributions inherent in rehearsal-free settings.
Extensive evaluations demonstrate that REAL-OW achieves state-of-the-art performance, surpassing existing exemplar replay methods in both detection precision and unknown discovery. Our approach establishes a new baseline for rehearsal-free OWOD. 

\end{abstract}

\section{Introduction}
\label{sec:intro}

Traditional object detection assumes a closed-set environment\cite{ren2016faster,carion2020end}, and it can only recognize categories that were defined during the initial training phase. However, real-world settings are dynamic and unpredictable\cite{scheirer2012toward}. To bridge this gap, Open-World Object Detection (OWOD) \cite{joseph2021towards} was introduced and it allows systems to operate in unconstrained settings\cite{gupta2022ow}. An OWOD detector must detect unknown objects and take them as novel classes, incrementally learning these new categories without forgetting previous ones \cite{wu2022uc,wang2023random,kirkpatrick2017overcoming,ma2023cat}. This ability is vital for computer vision scenarios like autonomous driving and open-ended robotics \cite{li2024open,joseph2022novel,ma2022rethinking}.

Despite its importance, current OWOD frameworks rely heavily on exemplar replay \cite{joseph2021towards, zohar2023prob, joseph2021incremental}, which mitigates catastrophic forgetting by maintaining a buffer of representative samples from previously learned classes and replaying them during new task training to preserve old knowledge\cite{rebuffi2017icarl,castro2018end,shmelkov2017incremental,hou2019learning}.

\begin{figure}[tb]
  \centering
  \includegraphics[height=6.8cm]{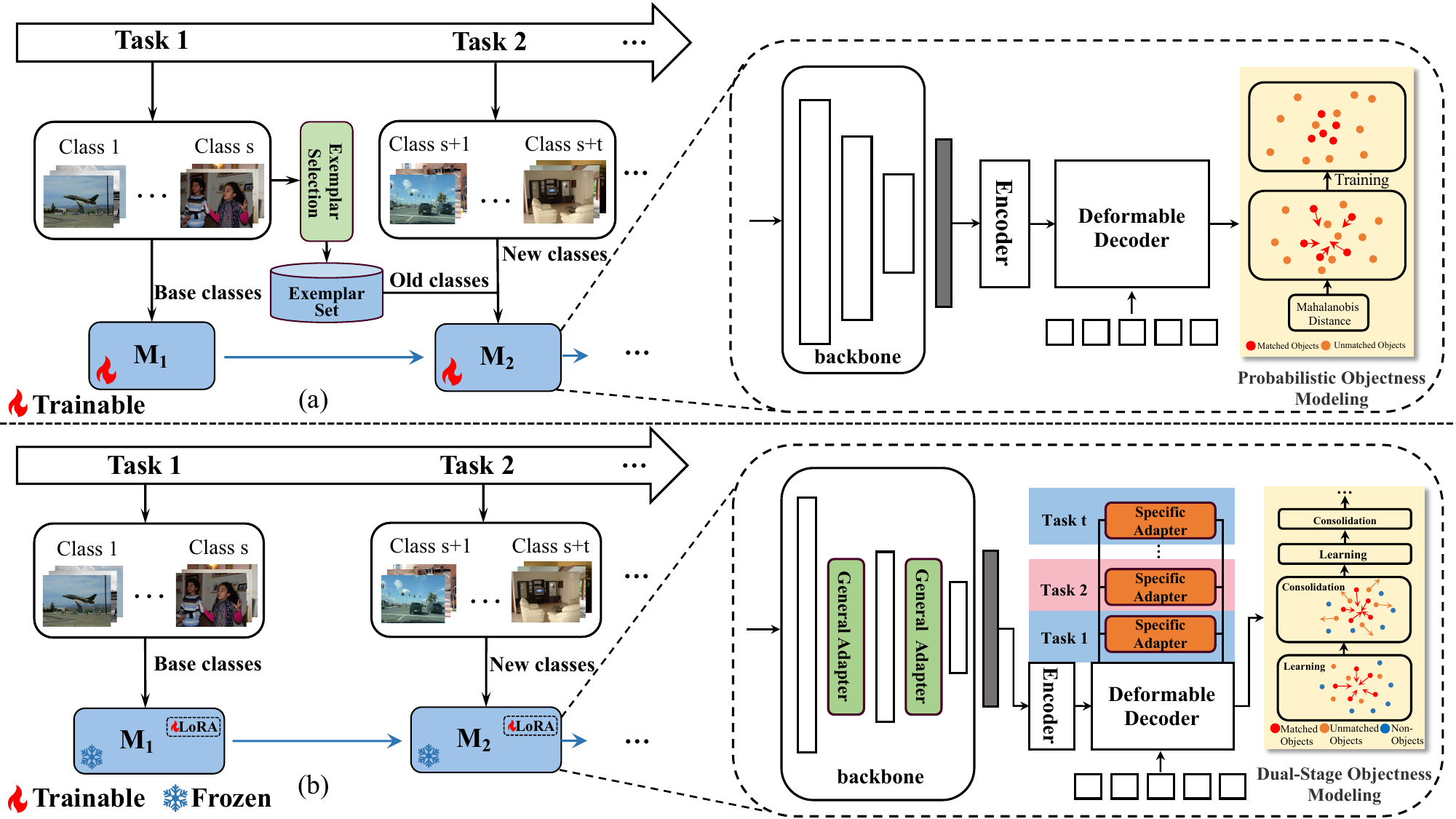}
  \caption{\textbf{Training process and architecture overview of our rehearsal-free framework.} We remove the exemplar replay mechanism from the typical training process in (a), achieving a more secure and streamlined rehearsal-free architecture design as shown in (b).
  To resist catastrophic forgetting without data replay, we employ a collaborative adapter architecture to separately capture general and task-specific features. Our Dual-Stage Objectness Modeling alternates between learning and consolidation phases,
  preventing representation drift that typically occurs in traditional probabilistic objectness modeling under rehearsal-free constraints.
  }
  \label{fig:intro}
\vspace{-5mm}
\end{figure}
However, exemplar replay incurs several practical challenges in real-world applications.
First, privacy-sensitive fields like medical imaging prevent models from retaining raw data after a task ends to avoid data exposure. This makes it prohibitive to store samples for exemplar replay\cite{prabhu2020gdumb}.
Second, as the task sequence expands, the buffer must accumulate more images to cover every previous class, thus incurring growing storage costs \cite{ermis2022memory}.
Furthermore, from the perspective of the OWOD objective, which is the continuous exploration and discovery of unknown categories, the learning process should prioritize modeling novel classes rather than repeatedly revisiting previously learned ones\cite{joseph2021towards,li2024open}.

These concerns and limitations motivate us to explore a rehearsal-free approach that avoids storing raw exemplar buffers. Low-Rank Adaptation (LoRA) \cite{hu2022lora} emerges as a compelling solution for its underlying parameter-efficient fine-tuning mechanism \cite{zhang2025c,lu2024adaptive,guo2024pilora,lin2024tracking,qiao2024learn}.
By constraining each added task updates into low-rank subspaces, LoRA enables the model to acquire new category knowledge while minimizing interference with previously learned representations \cite{xue2024adapting,zhang2025lori,qiao2024learn,wang2025smolora,wang2025malora}, and most importantly, it achieves this without relying on data replay.

In rehearsal-free OWOD, learning and preserving a continuous stream of knowledge without accessing old data is a fundamental challenge. A single shared representation cannot distinguish task-specific features, while completely independent modules hinder knowledge accumulation and transfer, leading to catastrophic forgetting\cite{zhang2025c,guo2024pilora}. To address this, we decouple general and specific knowledge by adopting a collaborative adapter framework, which separates general representations from task-specific expertise.
As illustrated in \cref{fig:intro}, our training process does not include any data replay. This avoids repeated data access, significantly enhancing data privacy protection and reducing storage requirements.
To balance general knowledge and task-specific expertise, our architecture employs two collaborative adapters within the frozen pre-trained model. Specifically, General Adapters (GAs) are placed in the backbone to maintain cross-task universal representations, while Specific Adapters (SAs) are added to the transformer decoder incrementally for each new task to capture knowledge for newly introduced categories.

Beyond incremental learning, OWOD further requires the discovery of unknown objects, which are not annotated but need to be discovered during training\cite{scheirer2012toward,joseph2021towards,li2024open}.
Recent methods employ probabilistic objectness\cite{zohar2023prob,wang2023orthogonal,sun2024exploring,doan2024hyp} and energy-based modeling\cite{energyout,zhang2025open} to discover unknown objects in unmatched queries and separate them from known classes and background, respectively.
However, these strategies depend on exemplar replay to stay distribution stable\cite{li2024open}. In rehearsal-free settings, the model continuously updates its parameters with emerging new data without access to historical samples, leading to potential feature drift, which causes two major problems:
(1) unknown decision bias, where emerging unknown class features are pulled toward current known class regions, causing them to inherit known class similarity bias, while unknown objects with significantly different textures from current known categories are often overly suppressed and misclassified as background;
and (2) catastrophic forgetting.

To solve these issues, we develop a Dual-Stage Objectness Modeling (DSOM) strategy. As shown in \cref{fig:intro}, this approach uses learning phases to aggregate the known feature embeddings and consolidation phases to sharpen the boundary between the knowns and unknowns.
Furthermore, we observe that under this rehearsal-free condition, the dual-stage update disperses the known object distribution, rendering the standard Mahalanobis distance\cite{zohar2023prob,zhang2025open,mashi} ineffective for uncertainty estimation due to its sensitivity to distribution compactness. To address this, we introduce a Calibrated Gaussian Negative Log-Likelihood (CG-NLL) distance, which accounts for the increased variance, better optimizes the expanded objectness density and stabilizes the energy landscape.

Experimental results demonstrate that our framework outperforms all compared exemplar replay methods in both precision and unknown discovery. Remarkably, our model achieves these results without replaying a single image. To the best of our knowledge, we are the first to propose a strictly rehearsal-free methodology for OWOD, thereby establishing a new baseline for the field.
Our main contributions can be summarized as follows:
\begin{itemize}
    \item We design General and Specific Adapters to continuously refine shared representations while isolating task-specific knowledge, enabling rehearsal-free open-world object detection without storing historical samples.

    \item We propose a Dual-Stage Objectness Modeling to avoid catastrophic forgetting caused by representation drift, together with a CG-NLL distance to calibrate objectness scores and stabilize the energy landscape.

    \item Extensive evaluations demonstrate that our approach outperforms all compared exemplar replay based methods, establishing a new and efficient baseline for rehearsal-free open-world object detection.
\end{itemize}

\section{Related Works}

\subsection{Open World Object Detection}
Open-World Object Detection (OWOD) shifts the focus from static closed-set recognition to dynamic environment adaptation\cite{scheirer2012toward,joseph2021towards}. In this paradigm, a detector must not only identify known categories but also detect previously unseen objects as unknown, subsequently incorporating them into the known set as new labels emerge\cite{ma2022rethinking,joseph2021incremental,li2024open}.
Joseph et al.\cite{joseph2021towards} formally introduced the OWOD framework by synthesizing concepts from Open Set Recognition\cite{scheirer2012toward} and Incremental Object Detection\cite{joseph2021incremental}. To alleviate catastrophic forgetting, they employed an exemplar replay (ER) mechanism, where a set of representative samples from earlier tasks is stored and replayed during incremental updates. Many follow-up approaches, such as OW-DETR\cite{gupta2022ow}, PROB\cite{zohar2023prob}, CAT\cite{ma2023cat}, and ORTH\cite{wang2023orthogonal}, continue to rely on exemplar buffers to preserve performance on prior classes during knowledge expansion.

However, storing raw data raises privacy and storage concerns like we mention above. Since current methods have not fully eliminated this dependency, a truly rehearsal-free approach is necessary.

\subsection{Parameter-Efficient Fine-Tuning}

Parameter-Efficient Fine-Tuning (PEFT) adapts large-scale models by updating only a small subset of parameters\cite{qiao2024learn,wang2023customizable}. Among these techniques, Low-Rank Adaptation (LoRA)\cite{hu2022lora} is a prominent
approach\cite{agiza2024mtlora,guo2024pilora,lin2024tracking,gandikota2024concept,zhang2025lori,valipour2023dylora,chavan2023one}. Formally, LoRA freezes the pre-trained weight matrix $W_0 \in \mathbb{R}^{d \times k}$ and introduces a low-rank update $\Delta W = BA$, where $B \in \mathbb{R}^{d \times r} $ and $A \in \mathbb{R}^{r \times k}$ are trainable matrices with rank $r \ll \min(d,k)$. This strategy significantly reduces the parameter space while preserving the generalization capability of the backbone.
Recent research\cite{zhang2023adalora,qiao2024learn,pu2025low,edalati2025krona}
focuses on enhancing LoRA through structural constraints and decoupled modularity.
In object detection,
some works\cite{chen2022vision,jia2022visual,pu2025low} use LoRA for domain-specific tasks, its potential for rehearsal-free Open-World Object Detection remains largely unexplored. The need to balance universal feature refinement with task-specific isolation in a strictly non-exemplar setting presents a unique challenge for current PEFT designs in detection domains.

\subsection{Class-agnostic Objectness Modeling}

In OWOD, class-agnostic objectness modeling decouples object existence from category identity by estimating whether a region corresponds to any valid object instance, thereby enabling the discovery of previously unseen categories\cite{ren2016faster,zhang2025open,Tian_2019_ICCV, jaiswal2021class,maaz2022class}.
Early methods such as PROB\cite{zohar2023prob} and OrthogonalDet\cite{sun2024exploring} estimate objectness through probabilistic modeling and orthogonal decomposition. More recent works, such as PASS\cite{Yang_2025_CVPR}, leverage attribute subsets to refine proposal scoring, while OWOBJ\cite{zhang2025open} assumes a dynamic Gaussian prior over objectness features and scores proposals using Mahalanobis distance. However, the optimization of these methods is typically restricted to current task data, relying heavily on exemplar replay to prevent forgetting. Without data rehearsal, continuous model updates cause representation drift, which destabilizes the objectness modeling and leads to a collapse in novelty discovery.

\section{Methods}
REAL-OW is a decoupled framework that synergizes General Adapters and Specific Adapters with Dual-Stage Objectness Modeling (DSOM) for rehearsal-free OWOD. Built on D-DETR-based detectors\cite{carion2020end,gupta2022ow}, this architecture balances general and task-specific knowledge refinement for each incremental task with stable feature accumulation and robust unknown discovery. Together, these components enable incremental learning without relying on historical samples.

\begin{figure}[tb]
  \centering
  \includegraphics[height=6.8cm]{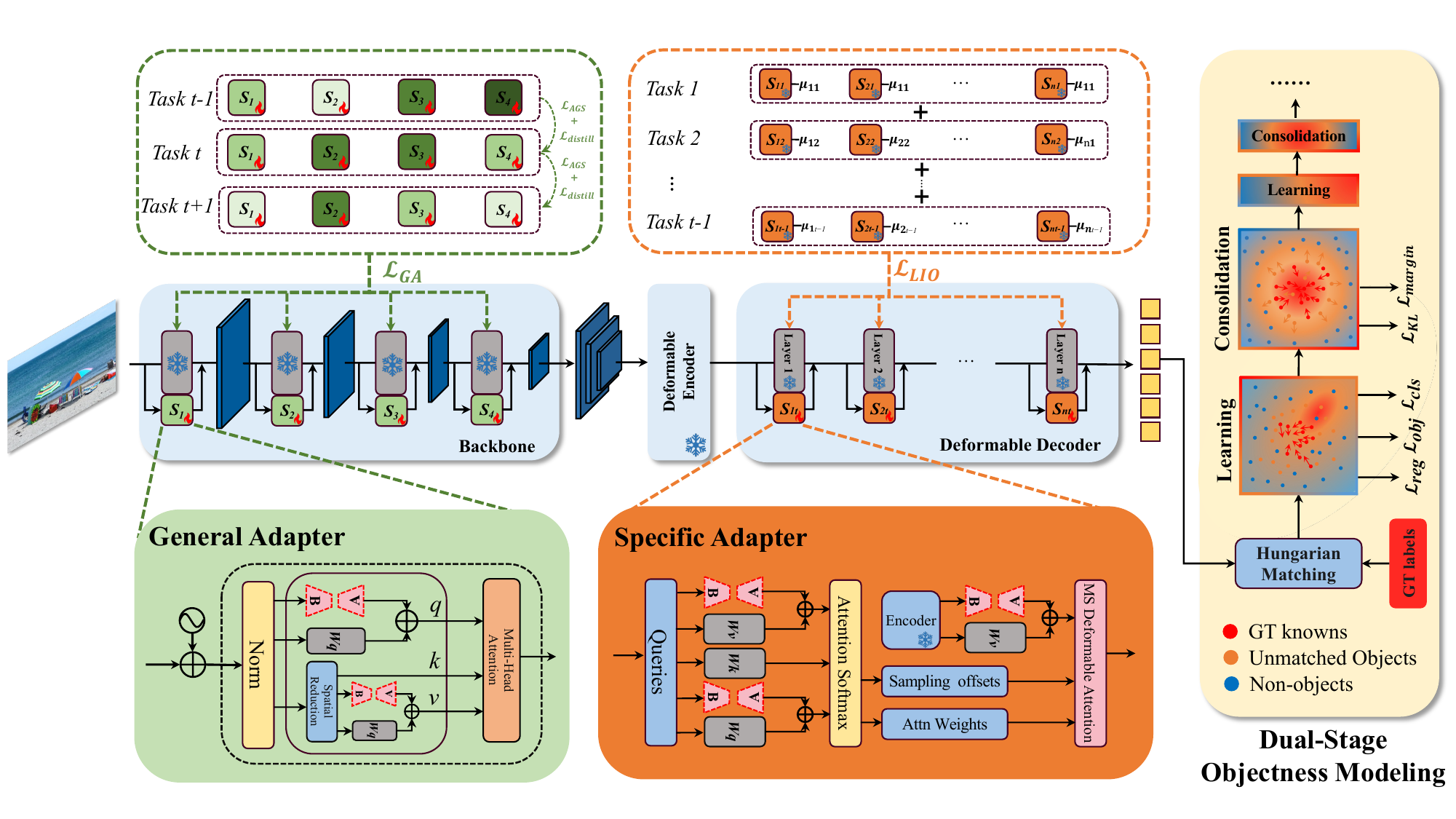}
  \caption{\textbf{Pipeline of our REAL-OW.} Our architecture consists of three core components: (1) \textbf{General Adapters} in the backbone, which employ the $\mathcal{L}_{AGS}$ to enable a significance-aware refinement of cross-task universal representations;  (2) \textbf{Specific Adapters} in the decoder layers, which are added incrementally to store task-unique features in non-overlapping subspaces; and (3) \textbf{Dual-Stage Objectness Modeling (DSOM).} The DSOM module alternates between a learning phase for feature aggregation and a consolidation phase for boundary refinement. This synergistic design enables the model to accumulate knowledge and discover novel objects without relying on historical samples.}
  \label{fig:pipeline}
\vspace{-5mm}
\end{figure}

\subsubsection{General Adapters.}

We adopt General Adapters to support the continuous refinement of shared representations within the Pyramid Vision Transformer (PVT) backbone. As illustrated in \cref{fig:pipeline}, we inject LoRA modules into the query $q$ and value $v$ projection matrices of the self-attention layers. During training, we keep the original backbone weights $W_{0}$ frozen and update only the lightweight GAs parameters. The forward pass for each GA can be formulated as:
\begin{equation}
f_{G}(x) = W_0 x + A_G B_G x,
\end{equation}
where x is input at any task $t$. $B_G \in \mathbb{R}^{r \times k}$ is a fixed, random orthogonal down-projection matrix, and $A_G \in \mathbb{R}^{d \times r}$ is a trainable up-projection matrix initialized to zero. This asymmetric design is motivated by recent work\cite{zhu2024asymmetrylowrankadaptersfoundation} showing that tuning $B$ is more impactful than tuning $A$.
To prevent standard regularization from indiscriminately suppressing all parameters through a uniform penalty, which leads to shrinkage bias\cite{wang2022learning,he2025cl}, we propose a novel Adaptive Gated Sparse Loss ($\mathcal{L}_{AGS}$) to optimize the GAs. This significance-aware mechanism allows the model to selectively accumulate impactful knowledge by dynamically modulating the regularization pressure. The loss function is formulated as:
\begin{equation}
\mathcal{L}_{AGS} = \sum_{\ell=1}^{L} {\left( \frac{1}{1 + \gamma \cdot \left( \|\mathbf{A}_\ell\|_F + \|\mathbf{B}_\ell\|_F \right)} \right)} \cdot {\left( \|\mathbf{A}_\ell\|_F + \|\mathbf{B}_\ell\|_F \right)},
\label{eq:ga}
\end{equation}
where $L$ is the number of backbone layers. The fractional term serves as an adaptive gate, which acts as a dynamic switch by sensing the Frobenius norm ($\|\cdot\|_F$) of each adapter layer. Specifically, the gate reduces regularization pressure for essential layers to preserve them as stable anchors for universal knowledge, and it applies higher sparse pressure to redundant layers, driving them to optimize, adapting and capturing shared features in later tasks of the sequence. To ensure high-fidelity consistency with the teacher model as well, we employ $\mathcal{L}_{distill}$, an output-aligned distillation loss, formulated as:
\begin{equation}
\mathcal{L}_{distill} = \sum_{i \in \mathcal{C}_t} s_{t-1,i} \log(s_{t,i})
\label{eq:distill},
\end{equation}
where $s_t$ represents the normalized feature distribution from backbone over task $t$ classes $\mathcal{C}_t$.
Combined with \cref{eq:ga} and \cref{eq:distill}, the whole GA loss formulated as:
\begin{equation}
\mathcal{L}_{GA} = \mathcal{L}_{distill} + \mathcal{L}_{AGS}.
\label{eq:fa}
\end{equation}

\subsubsection{Specific Adapters.}
While the GAs focuses on refining cross-task universal representations in the backbone, the feature interaction part requires specialized expertise to handle task-specific category features.
To this end, we employ the Specific Adapters in the decoder.
As shown in \cref{fig:pipeline}, the SAs employ the same LoRA placement in the $q$ and $v$ projections and the same asymmetric initialization strategy. But for each new task, a dedicated set of SA modules is added in parallel. This incremental expansion enables the decoder to store task-specific knowledge in isolated branches.

However, this multi-branch architecture is susceptible to feature entanglement. Because LoRA updates only a small fraction of the parameters relative to the pre-trained model, discriminating between unique task features remains difficult. Current incremental methods\cite{zhang2025c,qiao2024learn} often address this by using scalar control gates but these mechanisms are content-agnostic, as they ignore actual parameter magnitudes, making them insufficient to fully prevent interference between tasks.
To ensure that newly added adapters do not interfere with the specialized knowledge stored during preceding task stages, we optimize the SAs using the Layer-wise Interaction Orthogonal Loss ($\mathcal{L}_{LIO}$). Unlike standard geometric constraints, our approach is capacity-aware by coupling the gating signal with the actual parameter intensity. The loss is formulated as:
\begin{equation}
\mathcal{L}_{LIO} = \sum_{i=1}^{t-1} \sum_{k=1}^{N} \left( \mu_t^k \cdot \text{sg} \left[ \mu_i^k \cdot \|\mathbf{A}_i^k\|_F \right] \right)^2,
\label{eq:cga}
\end{equation}
where $N$ represents the number of layers in decoder. $\mu_t^k $ denotes the learnable weight for task $ t $ at the $ k $-th block, and $\|\mathbf{A}_i^k\|_F$ represents the Frobenius norm of the adapter parameters from previous task $ i $. The operator $ \text{sg}\left[  \cdot \right] $ indicates a stop-gradient operation to protect historical weights. The product $\mu_i^k \cdot \|\mathbf{A}_i^k\|_F$ defines the effective strength of a layer, which quantifies both its activation level and stored feature intensity.
$\mathcal{L}_{LIO}$ dynamically guides the current task to utilize layers that are either structurally inactivated or parametrically sparse in history. This design ensures that each adapter group selects a non-overlapping set of significant layers, effectively isolating task-specific knowledge and preventing the corruption of prior feature representations.

\subsubsection{Dual-Stage Objectness Modeling.}
Objectness probability and energy-based modeling are essential for identifying candidate proposals and distinguishing between known objects, unknown objects, and background.
In a rehearsal-free setting where replay is unavailable, feature drift occurs,
which destabilizes the modeling process. To address this, we propose Dual-Stage Objectness Modeling (DSOM), which partitions the optimization process into alternating Learning and Consolidation phases. To further improve objectness quantification, we introduce CG-NLL distance.

\begin{wrapfigure}{r}{0.5\textwidth}
  \centering
  \vspace{-5mm}
  \includegraphics[width=\linewidth]{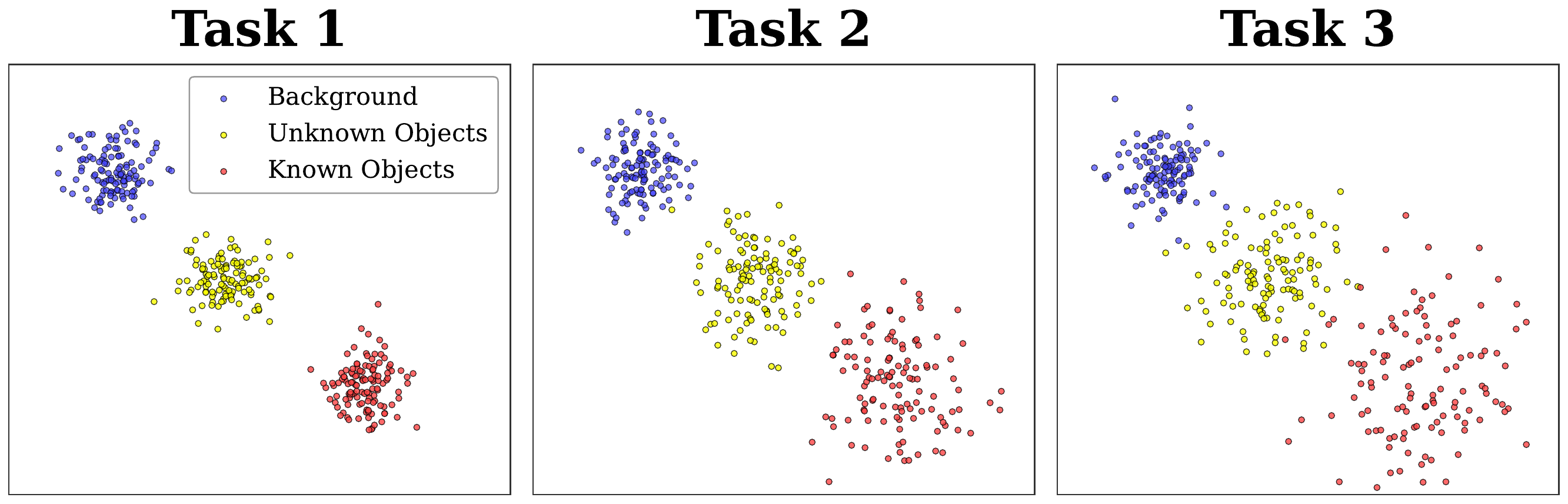}
  \caption{\textbf{t-SNE visualization of embeddings from Task 1 to Task 3}. Using the standard Mahalanobis distance without data replay, the known category distribution becomes progressively dispersed as tasks increase.
  }
  \label{fig:sne}
  \vspace{-8mm}
\end{wrapfigure}

\textbf{CG-NLL Distance.}
As shown in \cref{fig:sne},
we observe that known category distribution becomes more dispersed over time as semantic diversity grows without exemplar replay. Standard Mahalanobis distance fails to account for this expansion,
leading to inaccurate objectness estimation. To bridge the gap between
geometric distance and probabilistic objectness, we formulate the Calibrated Gaussian Negative Log-Likelihood (CG-NLL) distance to incorporate distribution density into a scaled objectness modeling. Our CG-NLL distance is defined as:
\begin{equation}
d_C(\mathbf{x}) = \frac{(\mathbf{x} - \boldsymbol{\mu})^\top \Sigma^{-1} (\mathbf{x} - \boldsymbol{\mu})}{1 + \beta \ln(1 + |\Sigma|)},
\label{eq:distance}
\end{equation}
where $\mu$ and $\Sigma$ represent the mean and covariance of the known category distribution.
We utilize the term $\tau(\Sigma) = 1 + \beta \ln(1 + |\Sigma|)$ as an adaptive scaling factor to normalize the distance based on the dispersion. $\beta$ is a calibration hyperparameter. By adaptively rescaling the Mahalanobis Distance,
CG-NLL distance maintains a strict manifold for compact distributions to suppress background noise, while providing a calibration for dispersed distributions to increase tolerance against catastrophic forgetting. Based on it, for unmatched queries $\boldsymbol{q}_i \in \mathcal{Q}_{\text{unk}}$, the objectness score
\begin{equation}
S_{obj}^i = \exp(-d_C(\mathbf{q}_i)),
\label{eq:score}
\end{equation}
serves as the confidence score for the unknown class pseudo-labels.

\textbf{Learning Phase.} This phase focuses on aggregating the feature representations of currently known categories and recognizes unknown objects. $\mathcal{L}_{Learning}$ is formulated as:
\begin{equation}
\mathcal{L}_{Learning} = \mathcal{L}_{obj} + \mathcal{L}_{cls} + \mathcal{L}_{reg},
\label{eq:l_learning}
\end{equation}
where $ \mathcal{L}_{obj} = \sum_{{q}_i \in \mathcal{Q_{\text{known}}}} d_C(q_i) $ minimizes the CG-NLL distance for matched known queries to improve objectness modeling, thereby enforcing a compact clustering of known features around the estimated distribution center.
$\mathcal{L}_{cls} = \text{FL}(\hat{l}, l)$ is the sigmoid focal loss for classification.
The predictions $\hat{l} \in \mathbb{R}^{Z+1}$ are supervised by targets $l$, which consist of discrete known class labels $l_{1:Z} \in \{0, 1\}^Z$ and a continuous objectness score $l_{Z+1} \in [0, 1]$. For unmatched queries, the $Z+1$-th dimension is assigned soft objectness pseudo-labels ${l}_i[Z+1] = S_{\text{obj}}^i$, given by \cref{eq:score}, allowing the model to regard high-scoring regions as potential unknown objects. $\mathcal{L}_{reg}$ is a combination of the $\ell_1$ loss and the GIoU loss.

\textbf{Consolidation Phase.} To maintain stability and sharpen the boundary between known and unknown energy distribution for preventing misclassification, the consolidation phase employs two optimization objectives:
\begin{equation}
\mathcal{L}_{Consolidation} = \mathcal{L}_{margin} + \mathcal{L}_{KL},
\label{eq:distance}
\end{equation}
where $\mathcal{L}_{margin}$ stands for energy margin loss, separating known and unknown energy distributions. $\mathcal{L}_{margin}$ is formulated as:
\begin{equation}
\mathcal{L}_{margin} = \mathbb{E}_{q^+ \in Q_{\text{known}}} \mathbb{E}_{q^- \in Q_{\text{unk}}} \left[ \max(0, E(q^+) - E(q^-) + \delta) \right],
\label{eq:distance}
\end{equation}
where $E(q) = -\log \sum_{i=1}^{C} \exp(f^i(q))$
denotes the energy level based on
the classifier logits $f^i(q)$. This hinge loss pushes the energy of unknown candidates above a margin $ \delta $ relative to knowns, establishing a discriminative surface. $\mathcal{L}_{KL}$ stands
for KL divergence loss\cite{zhang2025open}, acting as a
variance regularizer to prevent latent space collapse and maintaining embedding diversity, thereby mitigating over-fitting bias.
Through the alternation of these two phases, DSOM achieves a stable and rehearsal-free optimization of objectness knowledge across incremental tasks.

\section{Experiments}

\begin{table*}[t]
\centering
\caption{\textbf{Quantitative comparison for OWOD on M-OWODB (top) and S-OWODB (bottom).} Notably, our REAL-OW is \textbf{the only rehearsal-free approach} among all compared methods, yet it outperforms all other high-parameter CNN-based and D-DETR-based models that utilize exemplar replay. Best results are in \textbf{bold}.}
\label{tab:owod_traditional_methods}
\setlength{\tabcolsep}{2pt}
\adjustbox{width=\textwidth}{
\begin{tabular}{@{}l|cc|cccc|cccc|ccc@{}}
\toprule
 \textbf{Task IDs} ($\rightarrow$)& \multicolumn{2}{c|}{\textbf{Task 1}} & \multicolumn{4}{c|}{\textbf{Task 2}} & \multicolumn{4}{c|}{\textbf{Task 3}} & \multicolumn{3}{c}{\textbf{Task 4}} \\ \midrule

 & \cellcolor[HTML]{FFFFED}{U-Recall} & \multicolumn{1}{c|}{\cellcolor[HTML]{EDF6FF}{mAP ($\uparrow$)}} & \cellcolor[HTML]{FFFFED}{U-Recall} & \multicolumn{3}{c|}{\cellcolor[HTML]{EDF6FF}{mAP ($\uparrow$)}} & \cellcolor[HTML]{FFFFED}{U-Recall} & \multicolumn{3}{c|}{\cellcolor[HTML]{EDF6FF}{mAP ($\uparrow$)}} & \multicolumn{3}{c}{\cellcolor[HTML]{EDF6FF}{mAP ($\uparrow$)}}  \\

 & \cellcolor[HTML]{FFFFED}($\uparrow$) & \begin{tabular}[c]{@{}c}Current \\ known\end{tabular} & \cellcolor[HTML]{FFFFED}($\uparrow$) & \begin{tabular}[c]{@{}c@{}}Previously\\  known\end{tabular} & \begin{tabular}[c]{@{}c@{}}Current \\ known\end{tabular} & Both & \cellcolor[HTML]{FFFFED}($\uparrow$) & \begin{tabular}[c]{@{}c@{}}Previously \\ known\end{tabular} & \begin{tabular}[c]{@{}c@{}}Current \\ known\end{tabular} & Both & \begin{tabular}[c]{@{}c@{}}Previously \\ known\end{tabular} & \begin{tabular}[c]{@{}c@{}}Current \\ known\end{tabular} & Both \\

\midrule

UC-OWOD\cite{yu2023open} & \cellcolor[HTML]{FFFFED}2.4 & 50.7 & \cellcolor[HTML]{FFFFED}3.4 & 33.1 & 30.5 & 31.8 & \cellcolor[HTML]{FFFFED}8.7 & 28.8 & 16.3 & 24.6 & 25.6 & 15.9 & 23.2 \\
ORE\cite{joseph2021towards} & \cellcolor[HTML]{FFFFED}4.9 & 56.0 & \cellcolor[HTML]{FFFFED}2.9 & 52.7 & 26.0 & 39.4 & \cellcolor[HTML]{FFFFED}3.9 & 38.2 & 12.7 & 29.7 & 29.6 & 12.4 & 25.3 \\
2B-OCD\cite{wu2022two} & \cellcolor[HTML]{FFFFED}12.1 & 56.4 & \cellcolor[HTML]{FFFFED}9.4 & 51.6 & 25.3 & 38.5 & \cellcolor[HTML]{FFFFED}11.6 & 37.2 & 13.2 & 29.2 & 30.0 & 13.3 & 25.8 \\
OCPL\cite{maaz2022class} & \cellcolor[HTML]{FFFFED}8.3 & 56.6 & \cellcolor[HTML]{FFFFED}7.7 & 50.6 & 27.5 & 39.1 & \cellcolor[HTML]{FFFFED}11.9 & 38.7 & 14.7 & 30.7 & 30.7 & 14.4 & 26.7 \\
OW-DETR\cite{gupta2022ow} & \cellcolor[HTML]{FFFFED}7.5 & 59.2 & \cellcolor[HTML]{FFFFED}6.2 & 53.6 & 33.5 & 42.9 & \cellcolor[HTML]{FFFFED}5.7 & 38.3 & 15.8 & 30.8 & 31.4 & 17.1 & 27.8 \\
CAT\cite{ma2023cat} & \cellcolor[HTML]{FFFFED}23.7 & 60.0 & \cellcolor[HTML]{FFFFED}19.1 & 55.5 & 32.7 & 44.1 & \cellcolor[HTML]{FFFFED}24.4 & 42.8 & 18.7 & 34.8 & 34.4 & 16.6 & 29.9 \\
PROB\cite{zohar2023prob} & \cellcolor[HTML]{FFFFED}19.4 & 59.5 & \cellcolor[HTML]{FFFFED}17.4 & 55.7 & 32.2 & 44.0 & \cellcolor[HTML]{FFFFED}19.6 & 43.0 & 22.2 & 36.0 & 35.7 & 18.9 & 31.5 \\
OWOBJ\cite{zhang2025open} & \cellcolor[HTML]{FFFFED}23.6 & 61.4 & \cellcolor[HTML]{FFFFED}23.8 & {58.4} & 34.4 & 45.7 & \cellcolor[HTML]{FFFFED}25.1 & 44.8 & 27.8 & 38.8 & 36.4 & 20.7 & 32.0 \\
Hyp-OW\cite{doan2024hyp} & \cellcolor[HTML]{FFFFED}23.5 & 59.4 & \cellcolor[HTML]{FFFFED}20.6 & - & - & 44.0 & \cellcolor[HTML]{FFFFED}26.3 & - & - & 36.8 & - & - & 33.6 \\
RandBox\cite{wang2023random} & \cellcolor[HTML]{FFFFED}10.6 & {61.8} & \cellcolor[HTML]{FFFFED}6.3 & - & - & 45.3 & \cellcolor[HTML]{FFFFED}7.8 & - & - & 39.4 & - & - & 35.4 \\
ORTH\cite{sun2024exploring} & \cellcolor[HTML]{FFFFED} {24.6} & 61.3 & \cellcolor[HTML]{FFFFED} {26.3} & 55.5 & {38.5} & {47.0} & \cellcolor[HTML]{FFFFED} {29.1} & {46.7} & {30.6} & {41.3} & {42.4} & {24.3} & {37.9} \\
Ours:REAL-OW & \cellcolor[HTML]{FFFFED}\textbf{25.5} & \textbf{62.6} & \cellcolor[HTML]{FFFFED}\textbf{26.9} & \textbf{59.4} & \textbf{38.7} & \textbf{49.0} & \cellcolor[HTML]{FFFFED}\textbf{29.6} & \textbf{47.7} & \textbf{31.2} & \textbf{42.2} & \textbf{42.8} & \textbf{26.7} & \textbf{38.7} \\

\midrule

ORE\cite{joseph2021towards} & \cellcolor[HTML]{FFFFED}1.5 & 61.4 & \cellcolor[HTML]{FFFFED}3.9 & 56.5 & 26.1 & 40.6 & \cellcolor[HTML]{FFFFED}3.6 & 38.7 & 23.7 & 33.7 & 33.6 & 26.3 & 31.8 \\
OW-DETR\cite{gupta2022ow} & \cellcolor[HTML]{FFFFED}5.7 & 71.5 & \cellcolor[HTML]{FFFFED}6.2 & 62.8 & 27.5 & 43.8 & \cellcolor[HTML]{FFFFED}6.9 & 45.2 & 24.9 & 38.5 & 38.2 & 28.1 & 33.1 \\
PROB\cite{zohar2023prob} & \cellcolor[HTML]{FFFFED}17.6 & 73.4 & \cellcolor[HTML]{FFFFED}22.3 & 66.3 & 36.0 & 50.4 & \cellcolor[HTML]{FFFFED}24.8 & 47.8 & 30.4 & 42.0 & 42.6 & 31.7 & 39.9 \\
CAT\cite{ma2023cat} & \cellcolor[HTML]{FFFFED}24.0 & 74.2 & \cellcolor[HTML]{FFFFED}23.0 & 67.6 & 35.5 & 50.7 & \cellcolor[HTML]{FFFFED}24.6 & 51.2 & 32.6 & 45.0 & 45.4 & 35.1 & 42.8 \\
OWOBJ\cite{zhang2025open} & \cellcolor[HTML]{FFFFED}22.3 &{76.2} & \cellcolor[HTML]{FFFFED}{28.7} & {69.8} & {41.0} & {54.8} & \cellcolor[HTML]{FFFFED}30.9 & 50.6 & 35.7 & 46.8 & 46.7 & 36.9 & 43.2 \\
Hyp-OW\cite{doan2024hyp} & \cellcolor[HTML]{FFFFED}23.9 & 72.7 & \cellcolor[HTML]{FFFFED}23.3 & - & - & 50.6 & \cellcolor[HTML]{FFFFED}25.4 & - & - & 46.2 & - & - & 44.8 \\
ORTH\cite{sun2024exploring} & \cellcolor[HTML]{FFFFED}{24.6} & 71.6 & \cellcolor[HTML]{FFFFED}27.9 & 64.0 & 39.9 & 51.3 & \cellcolor[HTML]{FFFFED}{31.9} & {52.1} & {42.2} & {48.8} & {48.7} & {38.8} & {46.2} \\
Ours:REAL-OW & \cellcolor[HTML]{FFFFED}\textbf{24.9} & \textbf{76.9} & \cellcolor[HTML]{FFFFED}\textbf{29.8} & \textbf{71.5} & \textbf{42.3} & \textbf{56.9} & \cellcolor[HTML]{FFFFED}\textbf{33.3} & \textbf{53.4} & \textbf{44.6} & \textbf{50.5} & \textbf{49.7} & \textbf{39.3} & \textbf{47.1} \\

\bottomrule
\end{tabular}%
}
\vspace{0mm}
\end{table*}

\begin{table}[t]
\centering
\caption{\textbf{Comparison of trainable parameters.} Our method requires significantly fewer trainable parameters compared to existing OWOD paradigms.}
\label{tab:trainable_params}

\scriptsize

\renewcommand{\arraystretch}{1.2}
\setlength{\tabcolsep}{12pt}

\begin{tabular}{c|c}
\Xhline{1.2pt}
Methods & Trainable Parameters \\ \hline
Transformer-based & 39.7M \\
CNN-based         & 106.6M \\
Ours              & 2.0M \\
\Xhline{1.2pt}
\end{tabular}
\vspace{-4mm}
\end{table}

\begin{table*}[t]
\centering
\caption{Comparison for incremental object detection (iOD) on PASCAL VOC. We evaluate our model using three standard incremental scenarios. In these settings, new groups of classes (10, 5, or the final 1 class) are added to a detector that has already been trained on the initial set of classes (10, 15, or 19 classes, respectively). Best overall result for each setting are \textbf{bold}.}
\label{tab:iod}
\renewcommand{\arraystretch}{1.2}
\setlength{\tabcolsep}{2.5pt}

\resizebox{\textwidth}{!}{%
\begin{tabular}{l c c c || l c c c || l c c c}
\Xhline{1.2pt}
{\color[HTML]{009901} \textbf{10 + 10 setting}} & old cls. & new cls. & final mAP & {\color[HTML]{009901} \textbf{15 + 5 setting}} & old cls. & new cls. & final mAP & {\color[HTML]{009901} \textbf{19 + 1 setting}} & old cls. & new cls. & final mAP \\ \hline

ILOD\cite{shmelkov2017incremental}        & 63.2 & 63.1 & 63.2 & ILOD\cite{shmelkov2017incremental}        & 68.3 & 58.4 & 65.8 & ILOD\cite{shmelkov2017incremental}         & 68.5 & 62.7 & 68.2 \\
ORE\cite{joseph2021towards}          & 60.4 & 68.8 & 64.5 & ORE\cite{joseph2021towards}         & 71.8 & 58.7 & 68.5 & ORE\cite{joseph2021towards}          & 69.4 & 60.1 & 68.8 \\
OW-DETR\cite{gupta2022ow}      & 63.5 & 67.9 & 65.7 & OW-DETR\cite{gupta2022ow}      & 72.2 & 59.8 & 69.4 & OW-DETR\cite{gupta2022ow}      & 70.7 & 62.0 & 70.2 \\
PROB\cite{zohar2023prob}         & 66.0 & 67.2 & 66.5 & PROB\cite{zohar2023prob}         & 73.2 & 60.8 & 70.1 & PROB\cite{zohar2023prob}         & 73.9 & 48.5 & 72.6 \\
CAT\cite{ma2023cat}          & 67.9 & 67.4 & 67.7 & CAT\cite{ma2023cat}          & 75.6 & 59.3 & 72.2 & CAT\cite{ma2023cat}         & 74.5 & 61.1 & 73.8 \\
OWOBJ\cite{zhang2025open}        & 70.5 & 72.0 & 69.9 & OWOBJ\cite{zhang2025open}        & 76.5 & 63.7 & 73.3 & OWOBJ\cite{zhang2025open}        & 76.6 & 53.8 & 75.8 \\
ORTH\cite{sun2024exploring}         & 74.5 & 70.2 & 72.3 & ORTH\cite{sun2024exploring}        & 78.3 & 71.8 & 74.7 & ORTH\cite{sun2024exploring}       & 75.6 & 74.9 & 75.6 \\ \hline

\textbf{Ours:REAL-OW} & 73.8 & 74.3 & \textbf{74.0} & \textbf{Ours:REAL-OW} & 81.2 & 68.1 & \textbf{75.9} & \textbf{Ours:REAL-OW} & 78.2 & 60.7 & \textbf{77.3} \\
\Xhline{1.2pt}
\end{tabular}%
}
\vspace{-4mm}
\end{table*}

\begin{figure}[tb]
  \centering
  \includegraphics[width=\linewidth]{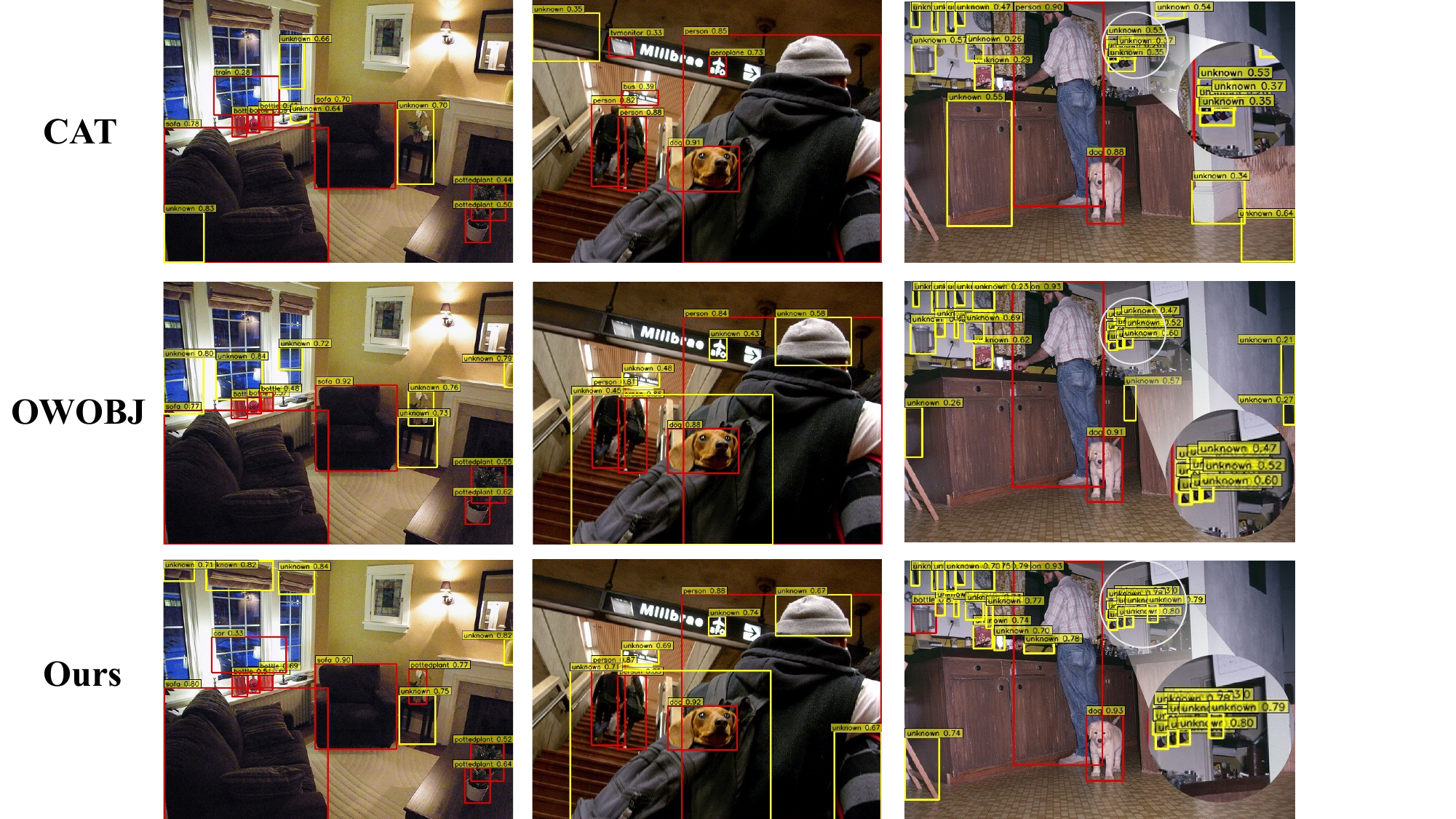}
  \caption{\textbf{Qualitative results comparison between CAT, OWOBJ and our REAL-OW.} Object detections with \textcolor{red}{Red}-known \textcolor{yellow}{Yellow}-unknown are displayed. Our method discovers more novel objects, such as the curtain in column 1 and the ladder in column 3. It also demonstrates more precise localization for multiple small objects, such as the cup cluster in column 3. In contrast, CAT misidentifies the tvmonitor in column 2, and OWOBJ mistakes a known pottedplant for an unknown object in column 1. These results show that our model provides better discovery and discrimination capabilities.
  }
  \label{fig:res_fig}
  \vspace{-5mm}
\end{figure}

\subsection{Experimental Setup}

\textbf{Datasets.} We evaluate REAL-OW on M-OWODB \cite{joseph2021towards} and S-OWODB \cite{gupta2022ow}, derived from MS-COCO \cite{lin2014microsoft} and PASCAL VOC \cite{everingham2010pascal}. S-OWODB groups classes by their super-categories like animals or vehicles, which is more challenging. Both benchmarks comprise four incremental tasks ($T_1$ to $T_4$), with 20 new classes introduced per task. During training, only labels for the current task are provided, while testing requires detecting all previously encountered categories. Most importantly, our approach is strictly rehearsal-free and stores no historical samples.
\\
\textbf{Evaluation Metrics.} We use mean Average Precision (mAP) to measure detection accuracy on known classes, and Unknown Recall (U-Recall) to evaluate the ability to discover unseen objects.
Additionally, A-OSE and WI are employed to quantify misclassifications between known and unknown categories.
\\
\textbf{Implementation Details.} We implement our framework in PyTorch and conduct all experiments on 3 NVIDIA RTX A$6000$ GPUs. Our model uses PVT as the backbone and Deformable DETR as the detection framework. Following standard practice, we pre-train the model on ImageNet \cite{deng2009imagenet} using a self-supervised approach to prevent premature exposure to future unknown classes \cite{gupta2022ow}. For a fair comparison, this pre-training setup is applied to all evaluated methods. The decoder consists of 6 layers. We set the ranks for the GAs and SAs to 16 and 10, respectively. Within the DSOM module, each iteration employs 2 learning cycles and 10 consolidation cycles. The hyperparameters $ \delta $, $\gamma$, $\beta$ are set to $0.2$, $0.01$, $0.1$ respectively. To ensure the reliability, all reported results represent the average performance calculated over five different random seeds.

\subsection{Experimental Results} \label{sec: exp}

For a fair comparison, we focus on models that generate unknown pseudo-labels internally, excluding approaches that leverage external large models, such as vision language models (VLMs),  to provide pseudo-label assignments.

\textbf{Quantitative.} As summarized in \cref{tab:owod_traditional_methods}, our REAL-OW framework achieves state-of-the-art performance on both M-OWODB and S-OWODB benchmarks. Notably, REAL-OW is the \textbf{only strictly rehearsal-free} method among all compared techniques. Despite this constraint, our model consistently achieves the top performance across all results, surpassing heavily parameterized CNN-based (e.g.RandBox, ORTH) and Transformer-based (e.g.CAT, PROB) methods. As shown in \cref{tab:trainable_params}, these baseline methods require over 50 times and nearly 20 times more trainable parameters than our approach, respectively. Furthermore, REAL-OW reaches the highest precision in Task 4, proving its superior resistance to catastrophic forgetting without any data replay. The performance gap between REAL-OW and OWOBJ also suggests that our dual-stage modeling is more effective than single-stage with known-unknown separation. REAL-OW also achieves the best results in A-OSE and WI among all compared methods. Refer to \cref{sec:metric2} in supplementary material for more details.

We evaluate REAL-OW on three standard iOD settings to assess knowledge retention. As shown in \cref{tab:iod}, our framework achieves the highest final mAP across all settings, outperforming SOTA methods. These results validate our architecture: GAs maintain stable feature foundations, while SAs and DSOM ensure task isolation and energy stability. Collectively, these components enable robust rehearsal-free learning with superior knowledge retention and exceptional resistance to catastrophic forgetting. Per-class details are in supplementary material \cref{sec:iod}.

\textbf{Qualitative.} As illustrated in \cref{fig:res_fig}, REAL-OW demonstrates higher U-Recall and more precise localization. It successfully discovers additional novel objects missed by other methods, such as the curtain in column $1$ and ladder in column $3$, while accurately separating individual cups in column $3$. These improvements show the CG-NLL distance effectively quantifies unknown objectness. Additionally, our model avoids the misclassification errors seen in CAT (tvmonitor, column $2$) and OWOBJ (pottedplant, column $1$). High confidence scores for these detections confirm that DSOM maintains a stable energy landscape for reliable boundary separation.
For more visualization results on the strong resistance of REAL-OW to catastrophic forgetting and its superior capability for discovering unknown classes, please refer to supplementary material \cref{sec:vis}.

\subsection{Ablation Study}

\textbf{LoRA.} We evaluate various LoRA rank configurations for the GAs and SAs. Following previous PEFT methods \cite{pu2025low,wang2025smolora,he2025cl}, we apply LoRA to $q$ and $v$ projections. Further validation for this placement provided in supplementary material \cref{sec:placement}. As shown in \cref{tab:rank}, we observe that performance does not scale linearly with the increase in trainable parameters. Instead, the configuration ($R_{GA}=16, R_{SA}=8$) achieves the optimal trade-off between performance and parameter capacity. Excessively high ranks lead to a decline in U-Recall and offer only marginal gains for mAP. This degradation is more pronounced as $R_{GA}$ increases, suggesting that an over-parameterized general feature space induces an unknown decision bias.
\begin{table*}[t]
\centering
\caption{\textbf{Results of different rank settings on Task 3.} We evaluate U-Recall and mAP across various rank combinations for GAs and SAs. The corresponding number of \textcolor{gray}{trainable parameters} is also reported.}
\label{tab:rank}
\resizebox{\textwidth}{!}{
    \setlength{\tabcolsep}{3pt}
    \renewcommand{\arraystretch}{1.1}
    \begin{tabular}{c|cc|cc|cc|cc|cc|cc} 
    \hline
    \multirow{2}{*}{\begin{tabular}[c]{@{}c@{}} GAs $\downarrow$ SAs $\rightarrow$ \\ (rank) \end{tabular}} & \multicolumn{2}{c|}{$r = 1$} & \multicolumn{2}{c|}{$r = 4$} & \multicolumn{2}{c|}{$r = 8$} & \multicolumn{2}{c|}{$r = 16$} & \multicolumn{2}{c|}{$r = 32$} & \multicolumn{2}{c}{$r = 64$} \\ \cline{2-13} 
     & U-Recall$\uparrow$ & mAP$\uparrow$ & U-Recall$\uparrow$ & mAP$\uparrow$ & U-Recall$\uparrow$ & mAP$\uparrow$ & U-Recall$\uparrow$ & mAP$\uparrow$ & U-Recall$\uparrow$ & mAP$\uparrow$ & U-Recall$\uparrow$ & mAP$\uparrow$ \\ \hline

    \multirow{2}{*}{$r = 1$} & $23.32$ & $35.31$ & $24.88$ & $38.69$ & $26.92$ & $40.37$ & $27.61$ & $40.44$ & $27.90$ & $39.23$ & $27.88$ & $38.84$ \\
      & \params{0.29} & \params{0.66} & \params{1.15} & \params{2.13} & \params{4.10} & \paramslast{8.03} \\ \hline 

    \multirow{2}{*}{$r = 4$} & $24.18$ & $36.28$ & $26.14$ & $39.60$ & $27.50$ & $41.86$ & $28.47$ & $41.75$ & $28.83$ & $41.02$ & $28.90$ & $40.54$ \\
      & \params{0.79} & \params{1.16} & \params{1.65} & \params{2.63} & \params{4.60} & \paramslast{8.53} \\ \hline

    \multirow{2}{*}{$r = 8$} & $24.98$ & $36.90$ & $27.03$ & $39.72$ & $28.82$ & $42.40$ & $28.90$ & $42.33$ & $28.85$ & $42.18$ & $28.87$ & $41.63$ \\
      & \params{1.45} & \params{1.82} & \params{2.31} & \params{3.30} & \params{5.26} & \paramslast{9.20} \\ \hline

    \multirow{2}{*}{$r = 16$} & $25.32$ & $37.78$ & $28.11$ & $40.80$ & $\mathbf{29.64}$ & $\mathbf{42.23}$ & $29.88$ & $42.11$ & $29.65$ & $41.20$ & $29.70$ & $40.75$ \\
       & \params{2.79} & \params{3.15} & \params{3.65} & \params{4.63} & \params{6.60} & \paramslast{10.53} \\ \hline

    \multirow{2}{*}{$r = 32$} & $24.64$ & $38.57$ & $28.46$ & $41.75$ & $28.70$ & $42.33$ & $29.42$ & $42.08$ & $29.48$ & $41.86$ & $29.60$ & $39.50$ \\
       & \params{5.45} & \params{5.82} & \params{6.31} & \params{7.29} & \params{9.26} & \paramslast{13.19} \\ \hline

    \multirow{2}{*}{$r = 64$} & $22.76$ & $38.61$ & $26.21$ & $41.81$ & $27.54$ & $42.21$ & $28.72$ & $41.97$ & $28.32$ & $40.22$ & $28.38$ & $38.43$ \\
       & \params{10.77} & \params{11.14} & \params{11.63} & \params{12.62} & \params{14.58} & \paramslast{18.51} \\ \hline
    \end{tabular}
}
\end{table*}

\begin{table}[t]
\centering
\caption{\textbf{\boldmath Performance comparison between standard distillation $\mathcal{L}_{distill}$ and our $\mathcal{L}_{AGS}$.} Best results are in \textbf{bold}.}
\label{tab:ablation_agsl}
\setlength{\tabcolsep}{2pt}
\adjustbox{width=\textwidth}{
\begin{tabular}{@{}l|cc|cccc|cccc|ccc@{}}
\toprule
 \textbf{Task IDs} ($\rightarrow$)& \multicolumn{2}{c|}{\textbf{Task 1}} & \multicolumn{4}{c|}{\textbf{Task 2}} & \multicolumn{4}{c|}{\textbf{Task 3}} & \multicolumn{3}{c}{\textbf{Task 4}} \\ \midrule

 & U-Recall & \multicolumn{1}{c|}{mAP ($\uparrow$)} & U-Recall & \multicolumn{3}{c|}{mAP ($\uparrow$)} & U-Recall & \multicolumn{3}{c|}{mAP ($\uparrow$)} & \multicolumn{3}{c}{mAP ($\uparrow$)}  \\

 & ($\uparrow$) & \begin{tabular}[c]{@{}c}Current \\ known\end{tabular} & ($\uparrow$) & \begin{tabular}[c]{@{}c@{}}Previously\\  known\end{tabular} & \begin{tabular}[c]{@{}c@{}}Current \\ known\end{tabular} & Both & ($\uparrow$) & \begin{tabular}[c]{@{}c@{}}Previously \\ known\end{tabular} & \begin{tabular}[c]{@{}c@{}}Current \\ known\end{tabular} & Both & \begin{tabular}[c]{@{}c@{}}Previously \\ known\end{tabular} & \begin{tabular}[c]{@{}c@{}}Current \\ known\end{tabular} & Both \\

\midrule

w/o $\mathcal{L}_{AGS}$ & \textbf{26.7} & \textbf{63.2} & 25.4 & 58.6 & \textbf{38.9} & 48.7 & 28.0 & 46.5 & 28.5 & 40.5 & 40.2 & 25.8 & 36.6 \\
w/ $\mathcal{L}_{AGS}$  & 25.5 & 62.6 & \textbf{26.9} & \textbf{59.4} & 38.7 & \textbf{49.0} & \textbf{29.6} & \textbf{47.7} & \textbf{31.2} & \textbf{42.2} & \textbf{42.8} & \textbf{26.7} & \textbf{38.7} \\

\bottomrule
\end{tabular}%
}
\end{table}

\begin{figure}[tb]
  \centering
  \includegraphics[width=\linewidth]{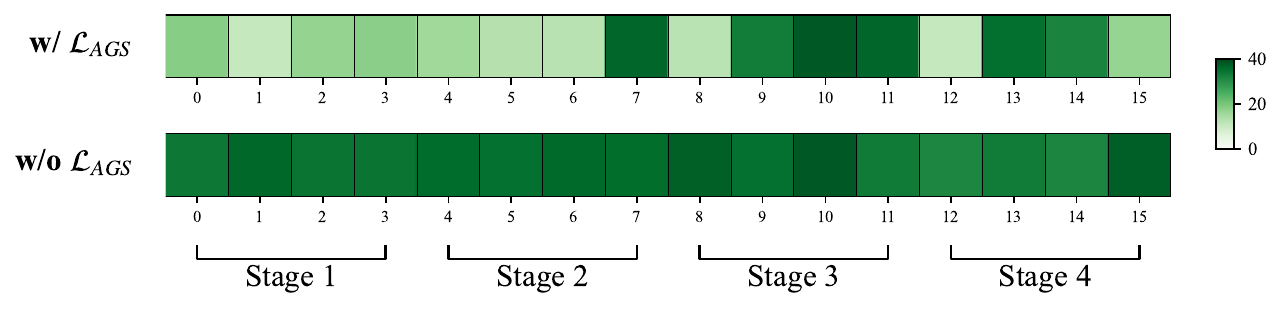}
  \caption{\textbf{Visualization of adapter weight magnitudes ($\ell_2$ norm) across backbone stages.} Deeper colors indicate higher layer significance. Our method (w/ $\mathcal{L}_{AGS}$) achieves a sparse and targeted update pattern, prioritizing significant layers while suppressing redundant ones.}
  \label{fig:weights}
\vspace{-3mm}
\end{figure}

\textbf{Adaptive Gated Sparse Loss.} \cref{tab:ablation_agsl} compares $\mathcal{L}_{AGS}$  with standard distillation $\mathcal{L}_{distill} $ that relies only on output alignment. As expected, during the initial task, $\mathcal{L}_{AGS}$ lags slightly behind the baseline due to its sparsity constraint.
However, by suppressing redundant layers, this constraint ensures that critical layers carry general features applicable to the encountered known categories, while other layers remain plastic. The performances on later tasks show the superior refinement of $\mathcal{L}_{AGS}$.
\cref{fig:weights} further confirms this targeted update pattern that critical layers preserve foundational representations with $\mathcal{L}_{AGS}$, while standard distillation leads to a uniform and undifferentiated weight distribution.

\begin{table}[!t]
\centering
\caption{\textbf{Ablation study of the Dual-Stage Objectness Modeling.} We report the results on Task 1 and Task 2. D-DETR upper limit denotes training with ground-truth unknown labels. D-DETR-ER uses standard exemplar replay, while D-DETR-RF employs our proposed GA and SA adapters for a rehearsal-free setting. Both baselines lack pseudo-labeling for unknown objects. We evaluate the impact of the CG-NLL distance, and the two training phases.}
\label{tab:ablation_DSOM}
\setlength{\aboverulesep}{0pt}
\setlength{\belowrulesep}{0pt}
\resizebox{\textwidth}{!}{
    \setlength{\tabcolsep}{6pt}
    \renewcommand{\arraystretch}{1.3}
    \begin{tabular}{l|cc|cccc}
    \hline
    \multirow{2}{*}{\textbf{Methods}} & \multicolumn{2}{c|}{\textbf{Task 1}} & \multicolumn{4}{c}{\textbf{Task 2}} \\ \cline{2-7}
     & U-Recall & mAP & U-Recall & Prev. & Curr. & Both \\ \hline

    D-DETR-upper limit & $32.9$ & $63.8$ & $40.5$ & $62.9$ & $41.2$ & $52.1$ \\
    D-DETR-ER & $-$ & $60.3$ & $-$ & $54.5$ & $34.4$ & $44.7$ \\
    D-DETR-RF & $-$ & $61.9$ & $-$ & $57.9$ & $38.5$ & $48.2$ \\ \midrule

    REAL-OW w/o CG-NLL  & $23.3$ & $58.3$ & $25.1$ & $57.5$ & $37.9$ & $47.7$ \\
    REAL-OW w/o Learning & $25.4$ & $60.5$ & $26.0$ & $47.8$ & $37.5$ & $42.7$ \\
    REAL-OW w/o Consolidation & $20.1$ & $63.1$ & $21.0$ & $58.9$ & $39.5$ & $49.2$ \\
    \textbf{REAL-OW - complete} & $\mathbf{25.5}$ & $\mathbf{62.6}$ & $\mathbf{26.9}$ & $\mathbf{59.4}$ & $\mathbf{38.7}$ & $\mathbf{49.0}$ \\ \hline
    \end{tabular}
}
\vspace{-5mm}
\end{table}

\begin{figure}[tb]
  \centering
  \includegraphics[trim={0mm 4mm 0mm 0mm}, clip, width=\linewidth]{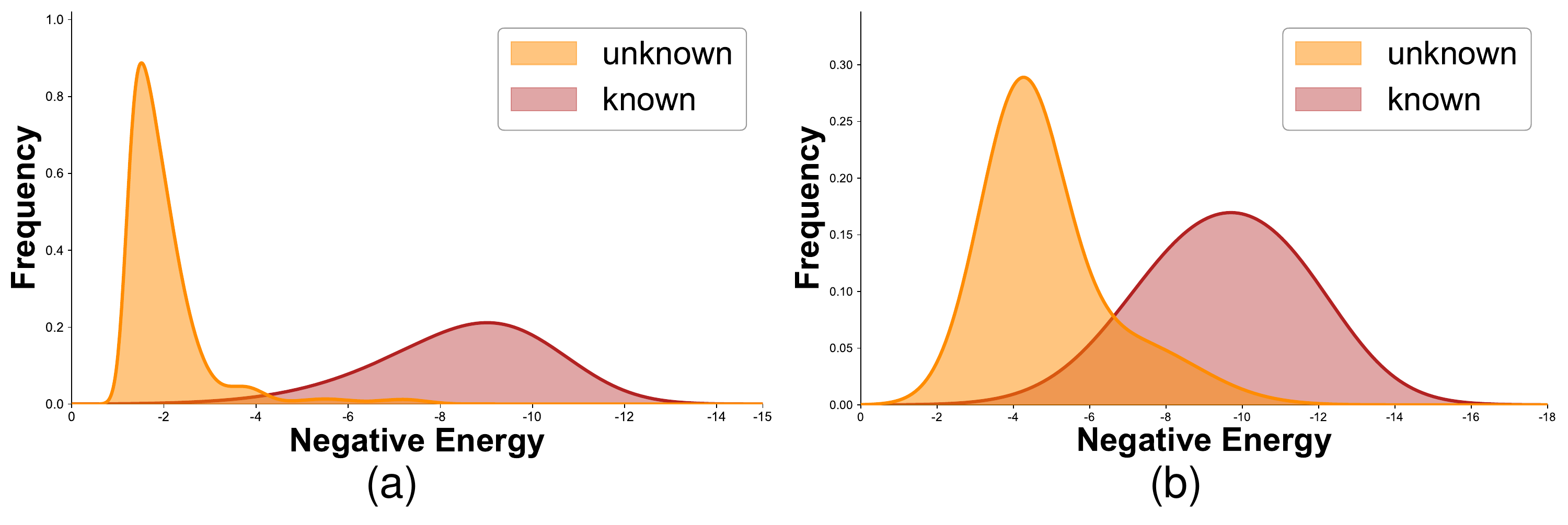}
  \caption{\textbf{Comparison between (a) CG-NLL distance and (b) standard Mahalanobis distance on negative energy distributions after Task 1.} CG-NLL distance effectively creates a clearer discriminative margin between known and unknown classes compared to the Mahalanobis distance.}
  \label{fig:cgnll}
\vspace{-6mm}
\end{figure}

\textbf{Dual-Stage Objectness Modeling.}
\cref{tab:ablation_DSOM} validates the core components of our DSOM.
\textbf{(1) CG-NLL Distance.}
Replacing CG-NLL distance with the standard Mahalanobis distance reduces U-Recall, demonstrating that standard metrics fail to calibrate energy scores as features disperse. This result is further explained in \cref{fig:cgnll}. By accounting for distribution dispersion, CG-NLL facilitates more accurate objectness modeling. This optimization enables the model to establish a significantly more distinct energy boundary between known and unknown classes.
\textbf{(2) The Learning and Consolidation Phases.}
Removing the learning phase causes previous mAP to drop from 59.4\% to 47.8\%, proving that feature aggregation is essential to prevent new knowledge from overwriting old representations. Conversely, skipping the consolidation phase slightly inflates current precision but collapses U-Recall from 26.9\% to 21.0\%. This indicates that without boundary separation, unknown objects are incorrectly biased toward known distributions and causes misclassification of the
known and unknown. 
\begin{wrapfigure}{r}{0.6\textwidth}
    \centering
    \vspace{0mm}
    \includegraphics[width=0.5\textwidth]{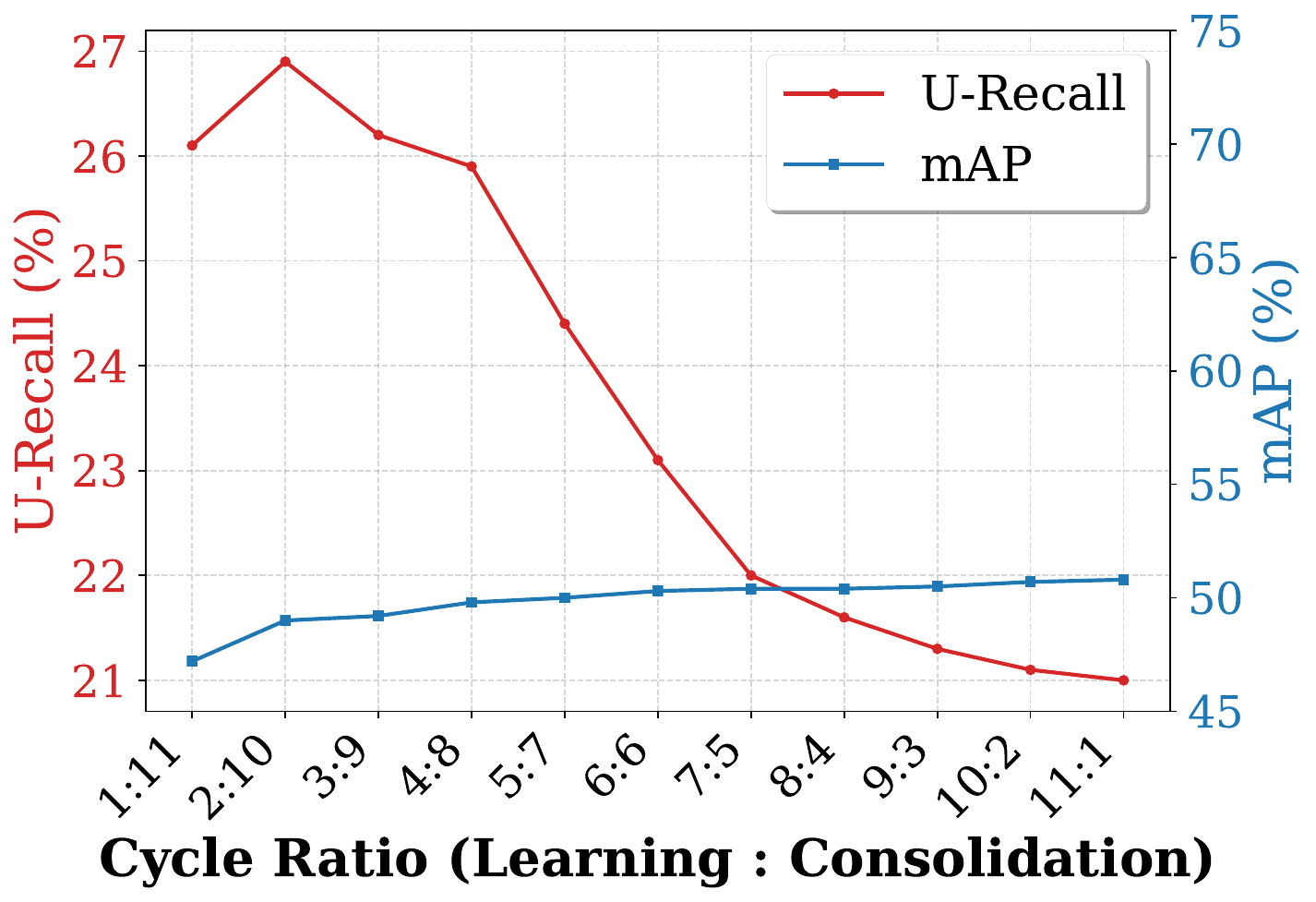}
    \vspace{-1mm}
    \caption{\textbf{Performance of different cycle ratios between learning and consolidation phases with total cycles fixed at 12.}}
    \label{fig:cycle}
    \vspace{-6mm}
\end{wrapfigure}
\textbf{(3) Cycle Ratios.} \cref{fig:cycle} illustrates the performance after Task $2$ under different cycle ratios of two phases. While mAP increases with more learning cycles, U-Recall peaks at a 2:10 ratio before declining. This confirms that while the learning phase maintains precision, allocating too many cycles to it creates a bias that hinders unknown discovery. We select 2:10 as the optimal balance to ensure both high stability and robust novelty detection.

\section{Conclusion}

In this paper, we propose REAL-OW, a novel rehearsal-free framework for Open-World Object Detection. REAL-OW decouples incremental knowledge through a collaborative adapter architecture, utilizing General Adapters for significance-aware feature refinement and Specific Adapters for orthogonal task isolation. To counteract representation drift, we introduce Dual-Stage Objectness Modeling (DSOM) supported by CG-NLL distance, which ensures robust objectness modeling and unknown discovery without data replay. Extensive experiments validate that REAL-OW outperforms all compared SOTA methods while maintaining a strictly rehearsal-free constraint.
The continuous stability of DSOM provides a reliable foundation for the future integration of external large models like vision-language models (VLMs), enabling the generation of high-quality pseudo-labels for future open-world research.

\section*{Acknowledgements}
This work was supported by the National Natural Science Foundation of China (Grant No. 62176163), the Shenzhen Higher Education Stable Support Program General Project (Grant No. 20231120175215001), the Scientific Foundation for Youth Scholars of Shenzhen University, the National Natural Science Foundation of China (Grant No. 62576218), the Guangdong Provincial Key Laboratory (Grant No. 2023B1212060076), and the Intelligent Computing Center of Shenzhen University.

\newcounter{mainfigcount}
\newcounter{maintabcount}
\setcounter{mainfigcount}{5}
\setcounter{maintabcount}{5}
\setcounter{figure}{\value{mainfigcount}}
\setcounter{table}{\value{maintabcount}}

\title{REAL-OW: \underline{RE}hears\underline{AL}-free \underline{O}pen \underline{W}orld Object Detection with Low-Rank Adaptation and Dual-Stage Objectness Modeling \\ Supplementary Material}
\titlerunning{REAL-OW}

\author{Huazhong Zhang\inst{1} \and
Yang Zhang\inst{1}\thanks{Corresponding author.} \and
Xiaowen Fu\inst{1} \and
Linlin Shen\inst{1} \and
Jinbao Wang\inst{1}
}

\authorrunning{H.~Zhang et al.}

\institute{Shenzhen University, Shenzhen, China\\
\email{yangzhang@szu.edu.cn}}

\maketitle

This supplementary material provides additional details and experimental evidence to support our main paper. We first present more detailed results for incremental object detection (iOD) and evaluate the model performance using the Wilderness Index (WI) and Absolute Open-Set Error (A-OSE). We also provide visualizations to demonstrate the forgetting resistance and discovery capabilities of REAL-OW. Finally, we include ablation studies to validate the placement of LoRA within attention projections.

\section{Incremental Object Detection} \label{sec:iod}

Previous works\cite{gupta2022ow,zohar2023prob} utilize incremental object detection (iOD) to evaluate model stability and knowledge retention. We assess the performance of REAL-OW on these iOD tasks, and the results are detailed in \cref{tab:iOD_detailed}. We compare our framework against current state-of-the-art methods, all of which rely on exemplar replay.
In each tested setting, the model is first trained on a base subset of categories (10, 15, or 19 classes), followed by incremental learning on the additional classes (10, 5, or 1 class).
As shown in \cref{tab:iOD_detailed}, our model achieves the highest mAP of 74.0, 75.9, and 77.3 across the three scenarios. Remarkably, REAL-OW outperforms these SOTA methods while remaining strictly rehearsal-free. These results demonstrate the exceptional robustness of our approach in incremental learning without any access to historical data.
This performance validates the effectiveness of our collaborative adapter architecture. In a rehearsal-free setting, General Adapters (GAs) update universal knowledge while Specific Adapters (SAs) preserve task-specific expertise. This collaborative design significantly mitigates catastrophic forgetting. Furthermore, the Dual-Stage Objectness Modeling (DSOM) phase establishes a stable and discriminative energy distribution, which allows the model to clearly separate known categories from unknown objects throughout the learning process.

\section{Performance on Metrics WI and A-OSE} \label{sec:metric2}

\cref{tab:owod_traditional_methods} in the main paper presents the performance of REAL-OW in mAP and U-Recall. To further analyze the detection quality, \cref{tab:metrics} provides a comparison using Wilderness Index (WI) and Absolute Open-Set Error (A-OSE). Specifically, WI measures the drop in precision for known categories when unknown objects are introduced into the evaluation. A-OSE counts the number of unknown instances that the model misclassifies as known classes. Together, these metrics quantify how often a model confuses unknown instances with background or known categories.
Notably, REAL-OW achieves the highest U-Recall while consistently maintaining the lowest WI and A-OSE across all task stages. These results indicate that our model effectively discovers novel objects without compromising the detection precision of known categories. Furthermore, it significantly reduces the misclassification of unknown instances as known classes. Such performance validates that the DSOM phase successfully establishes discriminative energy boundaries. Remarkably, REAL-OW maintains this robustness under strict rehearsal-free constraints, whereas most exemplar-replay-based methods struggle with error accumulation.

\setcounter{table}{6}
\begin{table}[t]
\vspace{0mm}
\centering
\caption{\textbf{State-of-the-art comparison for incremental object detection (iOD) on PASCAL VOC.} We experiment on 3 different settings. The comparison is shown in terms of per-class AP and overall mAP. The $10$, $5$ and $1$ class(es) in \colorbox{gray}{gray} background are introduced to a detector trained on the remaining $10$, $15$ and $19$ classes, respectively. }
\label{tab:iOD_detailed}
\adjustbox{width=\textwidth}{%
\begin{tabular}{@{}lccccccccccccccccccccc@{}}
\toprule
{\color[HTML]{009901} \textbf{10 + 10 setting}} & aero & cycle & bird & boat & bottle & bus & car & cat & chair & cow & table & dog & horse & bike & person & plant & sheep & sofa & train & tv & mAP \\ \midrule
ILOD\cite{shmelkov2017incremental}  & 69.9 & 70.4 & 69.4 & 54.3 & 48.0 & 68.7 & 78.9 & 68.4 & 45.5 & 58.1 & \cellcolor[HTML]{EEEEEE}59.7 & \cellcolor[HTML]{EEEEEE}72.7 & \cellcolor[HTML]{EEEEEE}73.5 & \cellcolor[HTML]{EEEEEE}73.2 & \cellcolor[HTML]{EEEEEE}66.3 & \cellcolor[HTML]{EEEEEE}29.5 & \cellcolor[HTML]{EEEEEE}63.4 & \cellcolor[HTML]{EEEEEE}61.6 & \cellcolor[HTML]{EEEEEE}69.3 & \cellcolor[HTML]{EEEEEE}62.2 & 63.2 \\
ORE\cite{joseph2021towards} & 63.5 & 70.9 & 58.9 & 42.9 & 34.1 & 76.2 & 80.7 & 76.3 & 34.1 & 66.1 & \cellcolor[HTML]{EEEEEE}56.1 & \cellcolor[HTML]{EEEEEE}70.4 & \cellcolor[HTML]{EEEEEE}80.2 & \cellcolor[HTML]{EEEEEE}72.3 & \cellcolor[HTML]{EEEEEE}81.8 & \cellcolor[HTML]{EEEEEE}42.7 & \cellcolor[HTML]{EEEEEE}71.6 & \cellcolor[HTML]{EEEEEE}68.1 & \cellcolor[HTML]{EEEEEE}77.0 & \cellcolor[HTML]{EEEEEE}67.7 & 64.5 \\
OW-DETR\cite{gupta2022ow} & 61.8 & 69.1 & 67.8 & 45.8 & 47.3 & 78.3 & 78.4 & 78.6 & 36.2 & 71.5 & \cellcolor[HTML]{EEEEEE}57.5 & \cellcolor[HTML]{EEEEEE}75.3 & \cellcolor[HTML]{EEEEEE}76.2 & \cellcolor[HTML]{EEEEEE}77.4 & \cellcolor[HTML]{EEEEEE}79.5 & \cellcolor[HTML]{EEEEEE}40.1 & \cellcolor[HTML]{EEEEEE}66.8 & \cellcolor[HTML]{EEEEEE}66.3 & \cellcolor[HTML]{EEEEEE}75.6 & \cellcolor[HTML]{EEEEEE}64.1 & 65.7 \\
PROB\cite{zohar2023prob} & 70.4 & 75.4 & 67.3 & 48.1 & 55.9 & 73.5 & 78.5 & 75.4 & 42.8 & 72.2 & \cellcolor[HTML]{EEEEEE}64.2 & \cellcolor[HTML]{EEEEEE}73.8 & \cellcolor[HTML]{EEEEEE}76.0 & \cellcolor[HTML]{EEEEEE}74.8 & \cellcolor[HTML]{EEEEEE}75.3 & \cellcolor[HTML]{EEEEEE}40.2 & \cellcolor[HTML]{EEEEEE}66.2 & \cellcolor[HTML]{EEEEEE}73.3 & \cellcolor[HTML]{EEEEEE}64.4 & \cellcolor[HTML]{EEEEEE}64.0 & 66.5 \\
CAT\cite{ma2023cat} & 76.5 & 75.7 & 67.0 & 51.0 & 62.4 & 73.2 & 82.3 & 83.7 & 42.7 & 64.4 & \cellcolor[HTML]{EEEEEE}56.8 & \cellcolor[HTML]{EEEEEE}74.1 & \cellcolor[HTML]{EEEEEE}75.8 & \cellcolor[HTML]{EEEEEE}79.2 & \cellcolor[HTML]{EEEEEE}78.1 & \cellcolor[HTML]{EEEEEE}39.9 & \cellcolor[HTML]{EEEEEE}65.1 & \cellcolor[HTML]{EEEEEE}59.6 & \cellcolor[HTML]{EEEEEE}78.4 & \cellcolor[HTML]{EEEEEE}67.4 & 67.7 \\
OWOBJ\cite{zhang2025open} & 75.9 & 80.7 & 73.3 & 52.1 & 58.8 & 77.7 & 81.9 & 79.1 & 47.9 & 77.8 & \cellcolor[HTML]{EEEEEE}70.4 & \cellcolor[HTML]{EEEEEE}78.9 & \cellcolor[HTML]{EEEEEE}80.3 & \cellcolor[HTML]{EEEEEE}79.3 & \cellcolor[HTML]{EEEEEE}80.0 & \cellcolor[HTML]{EEEEEE}45.1 & \cellcolor[HTML]{EEEEEE}70.2 & \cellcolor[HTML]{EEEEEE}78.4 & \cellcolor[HTML]{EEEEEE}68.5 & \cellcolor[HTML]{EEEEEE}68.4 & 69.9 \\
ORTH\cite{sun2024exploring}  & 82.4 & 77.3 & 78.2 & 59.7 & 61.2 & 84.3 & 90.1 & 80.2 & 49.8 & 81.7 & \cellcolor[HTML]{EEEEEE}58.2 & \cellcolor[HTML]{EEEEEE}74.0 & \cellcolor[HTML]{EEEEEE}82.9 & \cellcolor[HTML]{EEEEEE}81.0 & \cellcolor[HTML]{EEEEEE}81.2 & \cellcolor[HTML]{EEEEEE}38.3 & \cellcolor[HTML]{EEEEEE}70.8 & \cellcolor[HTML]{EEEEEE}68.0 & \cellcolor[HTML]{EEEEEE}77.4 & \cellcolor[HTML]{EEEEEE}70.2 & 72.3 \\
 \midrule
Ours: REAL-OW & 78.3 & 79.5 & 74.4 & 55.5 & 62.7 & 88.8 & 87.3 & 81.0 & 51.3 & 79.2 & \cellcolor[HTML]{EEEEEE}75.4 & \cellcolor[HTML]{EEEEEE}77.5 & \cellcolor[HTML]{EEEEEE}82.4 & \cellcolor[HTML]{EEEEEE}81.5 & \cellcolor[HTML]{EEEEEE}80.6 & \cellcolor[HTML]{EEEEEE}48.7 & \cellcolor[HTML]{EEEEEE}71.0 & \cellcolor[HTML]{EEEEEE}79.2 & \cellcolor[HTML]{EEEEEE}75.5 & \cellcolor[HTML]{EEEEEE}70.8 & \textbf{74.0} \\ \midrule \midrule
{\color[HTML]{009901} \textbf{15 + 5 setting}} & aero & cycle & bird & boat & bottle & bus & car & cat & chair & cow & table & dog & horse & bike & person & plant & sheep & sofa & train & tv & mAP \\ \midrule
ILOD\cite{shmelkov2017incremental} & 70.5 & 79.2 & 68.8 & 59.1 & 53.2 & 75.4 & 79.4 & 78.8 & 46.6 & 59.4 & 59.0 & 75.8 & 71.8 & 78.6 & 69.6 & \cellcolor[HTML]{EEEEEE}33.7 & \cellcolor[HTML]{EEEEEE}61.5 & \cellcolor[HTML]{EEEEEE}63.1 & \cellcolor[HTML]{EEEEEE}71.7 & \cellcolor[HTML]{EEEEEE}62.2 & 65.8 \\
ORE\cite{joseph2021towards} & 75.4 & 81.0 & 67.1 & 51.9 & 55.7 & 77.2 & 85.6 & 81.7 & 46.1 & 76.2 & 55.4 & 76.7 & 86.2 & 78.5 & 82.1 & \cellcolor[HTML]{EEEEEE}32.8 & \cellcolor[HTML]{EEEEEE}63.6 & \cellcolor[HTML]{EEEEEE}54.7 & \cellcolor[HTML]{EEEEEE}77.7 & \cellcolor[HTML]{EEEEEE}64.6 & 68.5 \\
OW-DETR\cite{gupta2022ow} & 77.1 & 76.5 & 69.2 & 51.3 & 61.3 & 79.8 & 84.2 & 81.0 & 49.7 & 79.6 & 58.1 & 79.0 & 83.1 & 67.8 & 85.4 & \cellcolor[HTML]{EEEEEE}33.2 & \cellcolor[HTML]{EEEEEE}65.1 & \cellcolor[HTML]{EEEEEE}62.0 & \cellcolor[HTML]{EEEEEE}73.9 & \cellcolor[HTML]{EEEEEE}65.0 & 69.4 \\
PROB\cite{zohar2023prob} & 77.9 & 77.0 & 77.5 & 56.7 & 63.9 & 75.0 & 85.5 & 82.3 & 50.0 & 78.5 & 63.1 & 75.8 & 80.0 & 78.3 & 77.2 & \cellcolor[HTML]{EEEEEE}38.4 & \cellcolor[HTML]{EEEEEE}69.8 & \cellcolor[HTML]{EEEEEE}57.1 & \cellcolor[HTML]{EEEEEE}73.7 & \cellcolor[HTML]{EEEEEE}64.9 & 70.1 \\
CAT\cite{ma2023cat} & 75.3 & 81.0 & 84.4 & 64.5 & 56.6 & 74.4 & 84.1 & 86.6 & 53.0 & 70.1 & 72.4 & 83.4 & 85.5 & 81.6 & 81.0 & \cellcolor[HTML]{EEEEEE}32.0 & \cellcolor[HTML]{EEEEEE}58.6 & \cellcolor[HTML]{EEEEEE}60.7 & \cellcolor[HTML]{EEEEEE}81.6 & \cellcolor[HTML]{EEEEEE}63.5 & 72.2 \\
OWOBJ\cite{zhang2025open} & 82.8 & 80.0 & 82.4 & 60.1 & 68.0 & 79.9 & 90.0 & 86.4 & 54.4 & 83.1 & 64.2 & 77.3 & 85.1 & 80.3 & 80.1 & \cellcolor[HTML]{EEEEEE}42.1 & \cellcolor[HTML]{EEEEEE}73.2 & \cellcolor[HTML]{EEEEEE}61.8 & \cellcolor[HTML]{EEEEEE}77.9 & \cellcolor[HTML]{EEEEEE}68.7 & 73.3 \\
ORTH\cite{sun2024exploring} & 82.7 & 80.4 & 78.5 & 55.3 & 65.5 & 81.0 & 89.8 & 85.9 & 52.6 & 84.6 & 62.3 & 78.4 & 82.7 & 81.1 & 84.2 & \cellcolor[HTML]{EEEEEE}46.5 & \cellcolor[HTML]{EEEEEE}71.6 & \cellcolor[HTML]{EEEEEE}79.0 & \cellcolor[HTML]{EEEEEE}82.5 & \cellcolor[HTML]{EEEEEE}79.2 & 74.7 \\
 \midrule
Ours: REAL-OW & 82.2 & 81.5 & 84.6 & 57.8 & 67.5 & 83.3 & 92.3 & 88.8 & 56.5 & 85.0 & 63.1 & 78.3 & 86.2 & 83.2 & 87.8 & \cellcolor[HTML]{EEEEEE}45.0 & \cellcolor[HTML]{EEEEEE}72.2 & \cellcolor[HTML]{EEEEEE}65.9 & \cellcolor[HTML]{EEEEEE}85.5 & \cellcolor[HTML]{EEEEEE}72.1 & \textbf{75.9} \\ \midrule \midrule
{\color[HTML]{009901} \textbf{19 + 1 setting}} & aero & cycle & bird & boat & bottle & bus & car & cat & chair & cow & table & dog & horse & bike & person & plant & sheep & sofa & train & tv & mAP \\ \midrule
ILOD\cite{shmelkov2017incremental} & 69.4 & 79.3 & 69.5 & 57.4 & 45.4 & 78.4 & 79.1 & 80.5 & 45.7 & 76.3 & 64.8 & 77.2 & 80.8 & 77.5 & 70.1 & 42.3 & 67.5 & 64.4 & 76.7 & \cellcolor[HTML]{EEEEEE}62.7 & 68.2 \\
ORE\cite{joseph2021towards} & 67.3 & 76.8 & 60.0 & 48.4 & 58.8 & 81.1 & 86.5 & 75.8 & 41.5 & 79.6 & 54.6 & 72.8 & 85.9 & 81.7 & 82.4 & 44.8 & 75.8 & 68.2 & 75.7 & \cellcolor[HTML]{EEEEEE}60.1 & 68.8 \\
OW-DETR\cite{gupta2022ow} & 70.5 & 77.2 & 73.8 & 54.0 & 55.6 & 79.0 & 80.8 & 80.6 & 43.2 & 80.4 & 53.5 & 77.5 & 89.5 & 82.0 & 74.7 & 43.3 & 71.9 & 66.6 & 79.4 & \cellcolor[HTML]{EEEEEE}62.0 & 70.2 \\
PROB\cite{zohar2023prob} & 80.3 & 78.9 & 77.6 & 59.7 & 63.7 & 75.2 & 86.0 & 83.9 & 53.7 & 82.8 & 66.5 & 82.7 & 80.6 & 83.8 & 77.9 & 48.9 & 74.5 & 69.9 & 77.6 & \cellcolor[HTML]{EEEEEE}48.5 & 72.6 \\
CAT\cite{ma2023cat} & 86.0 & 85.8 & 78.8 & 65.3 & 61.3 & 71.4 & 84.8 & 84.8 & 52.9 & 78.4 & 71.6 & 82.7 & 83.8 & 81.2 & 80.7 & 43.7 & 75.9 & 58.5 & 85.2 & \cellcolor[HTML]{EEEEEE}61.1 & 73.8 \\
OWOBJ\cite{zhang2025open} & 86.1 & 83.9 & 83.4 & 62.9 & 65.9 & 79.9 & 90.6 & 87.3 & 56.9 & 86.5 & 70.3 & 85.9 & 84.7 & 86.9 & 81.6 & 51.9 & 78.8 & 73.5 & 80.7 & \cellcolor[HTML]{EEEEEE}53.8 & 75.8 \\
ORTH\cite{sun2024exploring} & 83.8 & 84.7 & 77.0 & 62.9 & 60.8 & 80.9 & 88.6 & 85.8 & 51.1 & 81.4 & 67.2 & 86.7 & 86.3 & 83.4 & 83.4 & 44.7 & 74.5 & 73.1 & 81.1 & \cellcolor[HTML]{EEEEEE}74.9 & 75.6 \\
 \midrule
Ours: REAL-OW & 85.5 & 82.1 & 84.0 & 63.7 & 66.6 & 80.4 & 90.4 & 86.9 & 58.2 & 88.3 & 70.9 & 85.9 & 80.5 & 85.7 & 83.4 & 53.2 & 79.0 & 72.9 & 81.5 & \cellcolor[HTML]{EEEEEE}60.7 & \textbf{77.3} \\
\bottomrule
\end{tabular}%
}
\end{table}

\begin{table}[!t]
\vspace{0mm}
\centering
\caption{\textbf{Comparison with state-of-the-art methods on M-OWODB dataset. }Three metrics are reported: \textbf{U-Recall} measures the model's ability to retrieve unknown objects; \textbf{WI} (Wilderness Index) quantifies the degree of known classes being misclassified as unknown; \textbf{A-OSE} (Absolute Open-Set Error) counts the absolute number of unknown objects falsely identified as known classes.}
\label{tab:metrics}
\setlength{\tabcolsep}{5pt}
\adjustbox{width=\textwidth}{
\begin{tabular}{@{}l|ccc|ccc|ccc@{}}
\toprule
\textbf{Task IDs} ($\rightarrow$) & \multicolumn{3}{c|}{\textbf{Task 1}} & \multicolumn{3}{c|}{\textbf{Task 2}} & \multicolumn{3}{c}{\textbf{Task 3}} \\ \midrule

 & \cellcolor[HTML]{FFFFED}U-Recall & \cellcolor[HTML]{EDF6FF}WI & \cellcolor[HTML]{EDF6FF}A-OSE & \cellcolor[HTML]{FFFFED}U-Recall & \cellcolor[HTML]{EDF6FF}WI & \cellcolor[HTML]{EDF6FF}A-OSE & \cellcolor[HTML]{FFFFED}U-Recall & \cellcolor[HTML]{EDF6FF}WI & \cellcolor[HTML]{EDF6FF}A-OSE \\

 & \cellcolor[HTML]{FFFFED}($\uparrow$) & \cellcolor[HTML]{EDF6FF}($\downarrow$) & \cellcolor[HTML]{EDF6FF}($\downarrow$) & \cellcolor[HTML]{FFFFED}($\uparrow$) & \cellcolor[HTML]{EDF6FF}($\downarrow$) & \cellcolor[HTML]{EDF6FF}($\downarrow$) & \cellcolor[HTML]{FFFFED}($\uparrow$) & \cellcolor[HTML]{EDF6FF}($\downarrow$) & \cellcolor[HTML]{EDF6FF}($\downarrow$) \\ \midrule

ORE\cite{joseph2021towards} & \cellcolor[HTML]{FFFFED}4.9 & \cellcolor[HTML]{EDF6FF}0.0621 & \cellcolor[HTML]{EDF6FF}10459 & \cellcolor[HTML]{FFFFED}2.9 & \cellcolor[HTML]{EDF6FF}0.282 & \cellcolor[HTML]{EDF6FF}10445 & \cellcolor[HTML]{FFFFED}3.9 & \cellcolor[HTML]{EDF6FF}0.0211 & \cellcolor[HTML]{EDF6FF}7990 \\

2B-OCD\cite{wu2022two} & \cellcolor[HTML]{FFFFED}12.1 & \cellcolor[HTML]{EDF6FF}0.0481 & \cellcolor[HTML]{EDF6FF}- & \cellcolor[HTML]{FFFFED}9.4 & \cellcolor[HTML]{EDF6FF}0.160 & \cellcolor[HTML]{EDF6FF}- & \cellcolor[HTML]{FFFFED}11.6 & \cellcolor[HTML]{EDF6FF}0.0137 & \cellcolor[HTML]{EDF6FF}- \\

OW-DETR\cite{gupta2022ow} & \cellcolor[HTML]{FFFFED}7.5 & \cellcolor[HTML]{EDF6FF}0.0571 & \cellcolor[HTML]{EDF6FF}10240 & \cellcolor[HTML]{FFFFED}6.2 & \cellcolor[HTML]{EDF6FF}0.0278 & \cellcolor[HTML]{EDF6FF}8441 & \cellcolor[HTML]{FFFFED}5.7 & \cellcolor[HTML]{EDF6FF}0.0156 & \cellcolor[HTML]{EDF6FF}6803 \\

OCPL\cite{maaz2022class} & \cellcolor[HTML]{FFFFED}8.3 & \cellcolor[HTML]{EDF6FF}0.0413 & \cellcolor[HTML]{EDF6FF}5670 & \cellcolor[HTML]{FFFFED}7.6 & \cellcolor[HTML]{EDF6FF}0.0220 & \cellcolor[HTML]{EDF6FF}5690 & \cellcolor[HTML]{FFFFED}11.9 & \cellcolor[HTML]{EDF6FF}0.0162 & \cellcolor[HTML]{EDF6FF}5166 \\

PROB\cite{zohar2023prob} & \cellcolor[HTML]{FFFFED}19.4 & \cellcolor[HTML]{EDF6FF}0.0569 & \cellcolor[HTML]{EDF6FF}5195 & \cellcolor[HTML]{FFFFED}17.4 & \cellcolor[HTML]{EDF6FF}0.0344 & \cellcolor[HTML]{EDF6FF}6452 & \cellcolor[HTML]{FFFFED}19.6 & \cellcolor[HTML]{EDF6FF}0.0151 & \cellcolor[HTML]{EDF6FF}2641 \\

OWOBJ\cite{zhang2025open} & \cellcolor[HTML]{FFFFED}23.6 & \cellcolor[HTML]{EDF6FF}0.0395 & \cellcolor[HTML]{EDF6FF}3102 & \cellcolor[HTML]{FFFFED}23.8 & \cellcolor[HTML]{EDF6FF}0.0215 & \cellcolor[HTML]{EDF6FF}4032 & \cellcolor[HTML]{FFFFED}25.1 & \cellcolor[HTML]{EDF6FF}{0.0085} & \cellcolor[HTML]{EDF6FF}821 \\

\midrule \midrule

\textbf{REAL-OW} & \cellcolor[HTML]{FFFFED}\textbf{25.5} & \cellcolor[HTML]{EDF6FF}\textbf{0.0307} & \cellcolor[HTML]{EDF6FF}\textbf{2885} & \cellcolor[HTML]{FFFFED}\textbf{26.9} & \cellcolor[HTML]{EDF6FF}\textbf{0.0198} & \cellcolor[HTML]{EDF6FF}\textbf{3289} & \cellcolor[HTML]{FFFFED}\textbf{29.6} & \cellcolor[HTML]{EDF6FF}\textbf{0.0079} & \cellcolor[HTML]{EDF6FF}\textbf{711} \\

\bottomrule
\end{tabular}%
}
\vspace{0mm}
\end{table}

\setcounter{figure}{7}
\begin{figure}[t]
  \vspace{0cm}
  \centering
  \includegraphics[width=\linewidth]{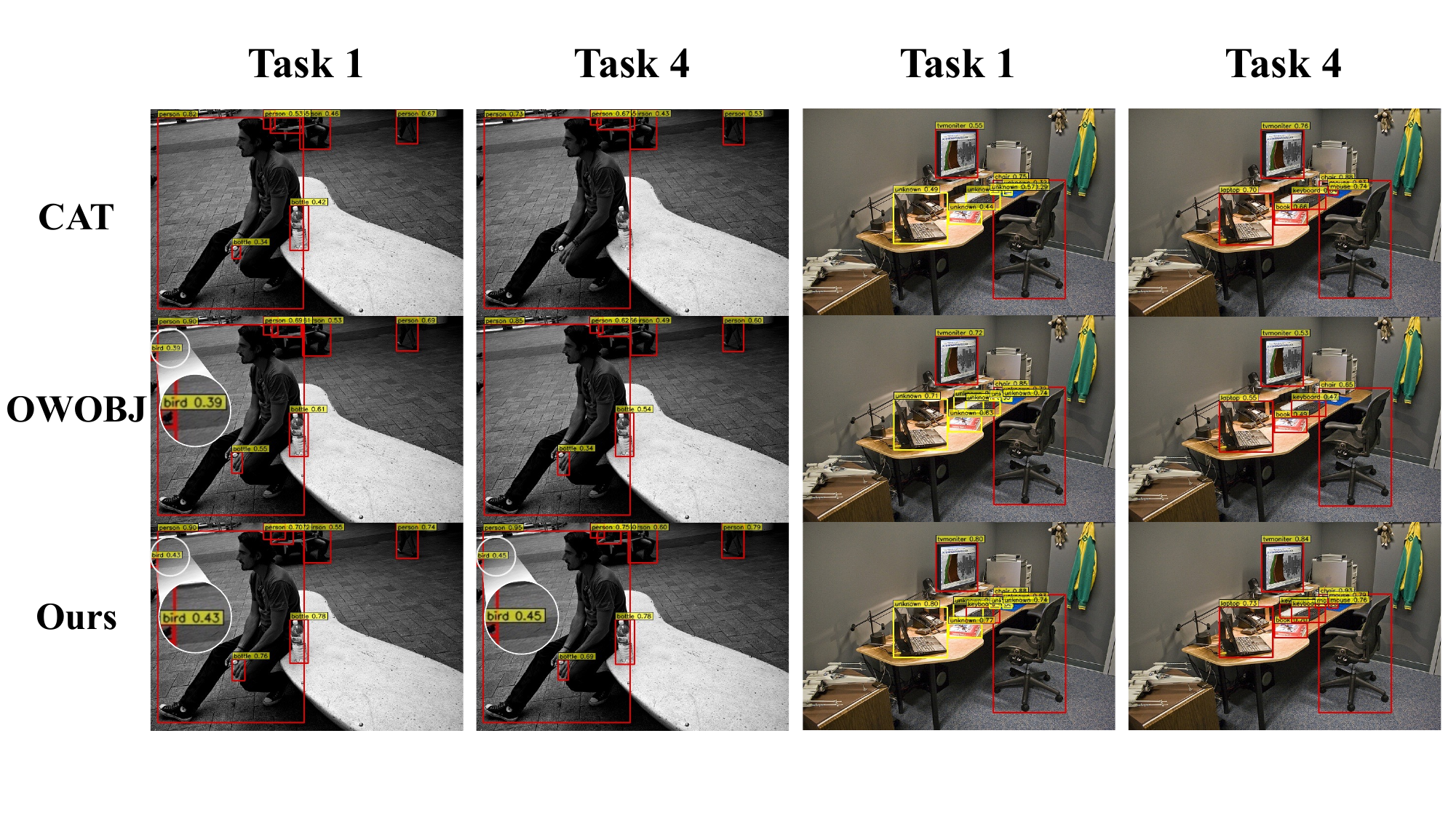}
  \vspace{-6mm}
  \caption{\textbf{Qualitative results comparison between CAT, OWOBJ and our REAL-OW.} Object detections with \textcolor{red}{Red}-known \textcolor{yellow}{Yellow}-unknown are displayed. Columns 1-2 demonstrate the resistance to catastrophic forgetting, where REAL-OW consistently identifies objects like the bottle and bird from Task 1 to Task 4. Columns 3-4 illustrate the process of learning unknown categories. REAL-OW labels the keyboard, mouse, and book as unknown in Task 1 and successfully learns them in Task 4, whereas competing methods suffer from missed detections or forget previously identified objects.
  }
  \label{fig:res_fig_supp}
  \vspace{0mm}
\end{figure}

\begin{table*}[t]
\vspace{0pt}
\centering
\caption{\textbf{Sensitivity analysis of adapter placement across representative rank regimes.} This table illustrates the interaction between LoRA placement and model capacity. The rows represent four placement configurations in the attention layers: $\{v\}$, $\{k, v\}$, $\{q, v\}$, and $\{q, k, v\}$. The columns denote various rank combinations for the General Adapter ($R_{GA}$) and Specific Adapter ($R_{SA}$). Within each cell, the top row provides the U-Recall (\%) and mAP (\%), while the bottom row indicates the total number of corresponding \textcolor{gray}{trainable parameters}.
}
\label{tab:position}
\resizebox{\textwidth}{!}{
    \setlength{\tabcolsep}{4pt}
    \renewcommand{\arraystretch}{1.3}
    \begin{tabular}{c|cc|cc|cc|cc|cc|cc|cc}
    \hline
    \multirow{2}{*}{\begin{tabular}{c} \textbf{$R_{GA}$:$R_{SA}$} $\to$ \\ \textbf{Attention Layers} $\downarrow$ \end{tabular}} & \multicolumn{2}{c|}{4 : 1} & \multicolumn{2}{c|}{4 : 4} & \multicolumn{2}{c|}{8 : 4} & \multicolumn{2}{c|}{8 : 8} & \multicolumn{2}{c|}{16 : 8} & \multicolumn{2}{c|}{32 : 16} & \multicolumn{2}{c}{64 : 32} \\ \cline{2-15}

    & U-Recall$\uparrow$ & mAP$\uparrow$ & U-Recall$\uparrow$ & mAP$\uparrow$ & U-Recall$\uparrow$ & mAP$\uparrow$ & U-Recall$\uparrow$ & mAP$\uparrow$ & U-Recall$\uparrow$ & mAP$\uparrow$ & U-Recall$\uparrow$ & mAP$\uparrow$ & U-Recall$\uparrow$ & mAP$\uparrow$ \\ \hline
    
    \multirow{2}{*}{$\{ v \}$} & 23.24 & 32.64 & 23.75 & 35.21 & 25.07 & 37.16 & 26.71 & 38.81 & 27.34 & 39.47 & 27.40 & 39.22 & 27.35 & 38.45 \\
    & \params{0.39} & \params{0.58} & \params{0.91} & \params{1.16} & \params{1.82} & \params{3.65} & \paramslast{7.29} \\ \hline

    \multirow{2}{*}{$\{k,v\}$} & 24.37 & 36.20 & 26.02 & 39.11 & 26.52 & 39.83 & 28.35 & 41.30 & 28.97 & 41.83 & 29.38 & 41.66 & 28.10 & 40.03 \\
    & \params{0.79} & \params{1.16} & \params{1.82} & \params{2.31} & \params{3.65} & \params{7.29} & \paramslast{14.58} \\ \hline

    \multirow{2}{*}{$\{q,v\}$} & 24.18 & 36.28 & 26.14 & 39.60 & 27.03 & 39.72 & 28.82 & 42.40 & \textbf{29.64} & \textbf{42.23} & 29.42 & 42.08 & 28.32 & 40.22 \\
    & \params{0.79} & \params{1.16} & \params{1.82} & \params{2.31} & \params{3.65} & \params{7.29} & \paramslast{14.58} \\ \hline

    \multirow{2}{*}{$\{q,k,v\}$} & 24.23 & 36.65 & 26.26 & 38.27 & 27.33 & 38.10 & 29.22 & 40.45 & 29.50 & 41.07 & 29.17 & 41.48 & 26.84 & 38.67 \\
    & \params{1.18} & \params{1.74} & \params{2.73} & \params{3.47} & \params{5.47} & \params{10.94} & \paramslast{21.87} \\ \hline
    \end{tabular}
}
\vspace{-4mm}
\end{table*}

\section{Visualization} \label{sec:vis}
We further illustrate the performance of REAL-OW in \cref{fig:res_fig_supp}, specifically focusing on its robustness in resisting catastrophic forgetting and its effectiveness in transitioning unknown objects into known categories in the task sequence.

\textbf{Resistance to Catastrophic Forgetting.} Columns 1 and 2 illustrate the performance of REAL-OW on long-term knowledge retention. For the initial classes identified in Task 1, competing methods show clear signs of forgetting by Task 4. Specifically, CAT fails to detect the bottle, and OWOBJ loses the bird. In contrast, REAL-OW persistently identifies these categories across incremental stages. This demonstrates significant resistance to catastrophic forgetting and maintains high localization precision without the need for data rehearsal.

\textbf{Learning Unknown Categories.} Columns 3 and 4 visualize the process of discovering novel objects and learning them incrementally. In Task 1, REAL-OW successfully identifies the book, keyboard, and mouse as unknown candidates. By Task 4, the model learns these objects as newly introduced known categories. Conversely, CAT fails to detect certain mouse and keyboard regions as unknowns in Task 1. Meanwhile, OWOBJ forgets its Task 1 unknown labels by the final stage, failing to recognize them as known classes in Task 4. These results confirm that our framework effectively discovers novel information and learns to recognize it in subsequent tasks.

\section{Ablation} \label{sec:placement}
\subsection{Ablation of LoRA Placement}
\vspace{-1mm}

Building on the analysis of \cref{tab:rank} in the main paper, we observe that optimal performance is concentrated in the asymmetric rank regime where $R_{GA} > R_{SA}$. This suggests that the backbone necessitates a higher parameter capacity to effectively refine and store universal representations, yielding a superior trade-off between accuracy and efficiency. Consequently, we fix the ranks within this optimal range to further investigate the impact of adapter placement within the Transformer architecture, as illustrated in \cref{tab:position}. According to these results, $\{q, v\}$ configuration remains the most robust across nearly all rank settings, validating its general superiority. We find that adding more adapter positions does not necessarily improve performance. Specifically, as the rank increases, using all three projections ($\{q, k, v\}$) yields lower accuracy than the $\{q, v\}$ or $\{k, v\}$ alternatives. This suggests that redundant projections introduce excessive task-specific noise and cause overfitting, which compromises stability and accelerates forgetting. Thus, the $\{q, v\}$ placement provides the most stable and parameter-efficient performance for our framework.

\begin{table}[!t]
\centering
\captionsetup{font=scriptsize, labelfont=bf}
\caption{\textbf{Ablation for loss components and $S_{obj}$ on M-OWODB.}}
\label{tab:tab_a}
\tiny
\setlength{\tabcolsep}{3.0pt}
\renewcommand{\arraystretch}{1.2}

\begin{tabular}{l|cc|cccc}

\hline
\multirow{2}{*}{\textbf{Methods}} & \multicolumn{2}{c|}{\textbf{Task 1}} & \multicolumn{4}{c}{\textbf{Task 2}} \\ \cline{2-7}
 & U-Recall & mAP & U-Recall & Prev. & Curr. & Both \\ \hline
w/o $S_{obj}$  & 10.8 & 55.7 & 11.2 & 53.1 & 33.9 & 43.5 \\
w/o $\mathcal{L}_{AGS}$ & 26.7 & 63.2 & 25.4 & 58.6 & 38.9 & 48.7 \\
w/o $\mathcal{L}_{distill}$ & 23.9 & 61.2 & 25.8 & 57.3 & 37.8 & 47.5 \\
w/o $\mathcal{L}_{LIO}$ & 24.5 & 62.2 & 23.6 & 55.8 & 37.6 & 46.7 \\
w/o $\mathcal{L}_{KL}$ & 20.4 & 61.9 & 21.8 & 57.5 & 38.0 & 47.7 \\
w/o $\mathcal{L}_{margin}$ & 20.8 & 62.3 & 22.3 & 58.0 & 37.3 & 47.6 \\
\hline
\textbf{REAL-OW (Ours)} & \textbf{25.5} & \textbf{62.6} & \textbf{26.9} & \textbf{59.4} & \textbf{38.7} & \textbf{49.0} \\
\hline
\end{tabular}
\vspace{-5mm}
\end{table}

\subsection{Ablation of main components}
\cref{tab:tab_a} presents a comprehensive ablation of REAL-OW on M-OWODB. Removing $S_{obj}$ causes Task 1 U-Recall to drop from 25.5\% to 10.8\%, underscoring its fundamental role in reliable unknown pseudo-labeling. Without $\mathcal{L}_{AGS}$, Task 2 performance degrades despite a marginal initial gain, confirming its importance in preserving universal representations for long-term stability. The exclusion of $\mathcal{L}_{LIO}$ results in a sharp Task 2 mAP drop from 49.0\% to 46.7\% due to severe feature entanglement, highlighting its critical function in enforcing task-specific isolation within the decoder. Finally, $\mathcal{L}_{KL}$ and $\mathcal{L}_{margin}$ jointly stabilize the latent objectness manifold and establish a discriminative energy margin to effectively separate known categories from unknowns.

\bibliographystyle{splncs04}
\bibliography{main}

\end{document}